\documentclass[10pt,journal,compsoc]{IEEEtran}
\ifCLASSOPTIONcompsoc
  \usepackage[nocompress]{cite}
\else
  \usepackage{cite}
\fi
\usepackage{graphicx}
\usepackage{amsmath,amssymb,amsfonts}
\usepackage{booktabs}
\usepackage{multirow}
\usepackage{array}

\usepackage{booktabs}

\usepackage{ragged2e}

\usepackage{multirow}

\usepackage[table,xcdraw,dvipsnames]{xcolor}
\usepackage{pifont}
\newcommand{\cmark}{\textcolor{green}{\ding{51}}}
\newcommand{\xmark}{\textcolor{red}{\ding{55}}}

\usepackage[normalem]{ulem}
\useunder{\uline}{\ul}{}

\ifCLASSINFOpdf
\else
\fi
\newcommand\MYhyperrefoptions{bookmarks=true,bookmarksnumbered=true,
pdfpagemode={UseOutlines},plainpages=false,pdfpagelabels=true,
colorlinks=true,linkcolor={black},citecolor={black},urlcolor={black},
pdftitle={FAST: Flow Any Scene Transformer},%<!CHANGE!
pdfsubject={ComputerVision},%<!CHANGE!
pdfauthor={Yongjian Zhang},%<!CHANGE!
pdfkeywords={Correspondence Matching, Foundation Model, Scaling Law}}%<^!CHANGE!
\usepackage[\MYhyperrefoptions,pdftex]{hyperref}
\begin{document}
\title{FAST: Flow Any Scene Transformer}

\author{Yongjian Zhang,
        Longguang Wang,
        Zhuo Song,
        Zhiheng Fu, % ~\IEEEmembership{Life~Fellow,~IEEE}
        Liang Lin~\IEEEmembership{Fellow,~IEEE},\\
        Yulan Guo$^*$~\IEEEmembership{Senior Member,~IEEE}% <-this % stops a space
\IEEEcompsocitemizethanks{
\IEEEcompsocthanksitem Yongjian Zhang, Longguang Wang, Zhuo Song and Yulan Guo are with the School of Electronics and Communication Engineering, the Shenzhen Campus of Sun Yat-sen University, Sun Yat-sen University, Shenzhen, China. 
Zhiheng Fu is with the Hong Kong Polytechnic University, HKSAR, China.
Liang Lin is with the School of Computer Science and Engineering, Sun Yat-sen University, Guangzhou, China. % \protect\\
% E-mail: zhangyj85@mail2.sysu.edu.cn, wanglongguang15@nudt.edu.cn, songzh@mail2.sysu.edu.cn, guoyulan@sysu.edu.cn 
\IEEEcompsocthanksitem $^*$Corresponding author: Yulan Guo. (guoyulan@mail.sysu.edu.cn)} % <-this % stops a space
}%\thanks{Manuscript received July 11, 2026.}}% ; revised August 26, 2015.}}

\IEEEtitleabstractindextext{%
\begin{abstract}
  \justifying
  Scaling has become a primary driver of progress in language and vision foundation models, yet its role in precise correspondence matching remains underexplored.
  In this work, we present Flow Any Scene Transformer (FAST), a scalable correspondence model driven by two key insights. 
  % First, rather than relying on data-constrained cross-view pretraining, we reveal that single-view pretrained ViTs trained at billion-sample scale provide a remarkably robust and scalable initialization for dense matching. 
  % Second, when transferring such backbones to two-view reasoning, explicit cross-attention is better aligned with the pretrained inductive biases than global self-attention.
  \textcolor{black}{
    % First, rather than relying on data-constrained cross-view pretraining, we reveal that the self-attention projections inside Vision Foundation Models encode a coarse yet reusable cross-view matching prior.
    % Second, to expose this prior for dense matching, rewiring self-attention into cross-attention provides a more favorable transfer initialization than global self-attention.}
    First, we reveal that the query-key projections inside single-view vision foundation models encode a coarse yet reusable prior for cross-view matching. 
    Second, reusing these pretrained projections in cross-attention form yields a highly effective initialization for a ViT-based matcher built from a single-view encoder.}
  Guided by these insights, we build FAST upon a vanilla single-view foundation model, utilizing a zero-parameter rewiring strategy to convert selected self-attention layers into cross-attention for cross-view interaction.
  % To fully unlock the scaling potential of this formulation, we further propose a depth-to-flow conversion pipeline that turns static rigid scenes into nearly 4 million optical-flow training pairs.
  \textcolor{black}{
    This design allows ViT-based matchers to scale with advances in single-view foundation models, bypassing the need for a dedicated pair-centric pretraining stage.
    To fully unlock the scaling potential of this formulation, we assemble a 6-million-pair training corpus for general-purpose dense 2D displacement estimation across diverse co-visible image pairs.}
  Extensive experiments demonstrate that FAST achieves state-of-the-art performance across a wide range of benchmarks, while scaling favorably with both backbone size and training data.
  % Instead of relying on task-specific architectural specialization, FAST is built on two simple design choices: reusing large-scale single-image pretrained backbones and fine-tuning on large-scale supervision. We demonstrate that a vanilla Vision Transformer pretrained on single-view images can serve as the matching network itself, with performance improving consistently as the backbone scales up, thereby eliminating the need for costly cross-view pretraining. Besides, we introduce depth-to-flow, a geometric conversion that turns stereo/MVS/SLAM collections of static rigid scenes into millions of optical-flow labels, thereby easily obtaining scalable fine-tuning data for learning correspondence-specific inductive bias. With these design choices, FAST achieves state-of-the-art performance on stereo matching, optical flow, and sparse matching benchmarks, and generalizes robustly to in-the-wild scenes.
\end{abstract}

\begin{IEEEkeywords}
  Correspondence matching, Foundation model, Scaling laws.
\end{IEEEkeywords}}

\maketitle

% is not selected <OR> if conference mode is selected - because compsoc
% conference papers position the abstract like regular (non-compsoc)
\IEEEdisplaynontitleabstractindextext

% \fi
\IEEEpeerreviewmaketitle

\ifCLASSOPTIONcompsoc
\IEEEraisesectionheading{\section{Introduction}\label{sec:introduction}}
\else
\section{Introduction}
\label{sec:introduction}
\fi

\IEEEPARstart{C}{orrespondence} matching is the cornerstone of many cross-view tasks, serving as a critical requirement for applications like reconstruction and navigation. 
\textcolor{black}{Over the past decades, this task has evolved into a standardized volume-centric pipeline~\cite{stereo_survey}, where view-wise representations are converted into a cost-defined hypothesis space and subsequent modules process over the resulting cost volume.
Although highly successful, this paradigm suffers a limited model scaling route, as simply stacking more aggregation or refinement blocks has shown diminishing gains~\cite{DCANet,SEA-RAFT}.
Further gains therefore rely heavily on improved module and architecture designs.}

\textcolor{black}{By contrast, recent ViT-based approaches~\cite{croco, VGGT} introduce a more scalable architectural paradigm, where visual representations and matching cues are progressively refined through stacked Transformer blocks.
However, scaling such models relies on pair-centric pretraining on structurally overlapping image pairs, which are not as easy to collect at scale as the independent images used by single-view ViTs.
For instance, CroCo~\cite{croco} is pretrained on 5 million pairs, while DINOv3~\cite{dinov3} employs 1689 million images.
At the same time, recent advances~\cite{FoundationStereo,RoMa,PanMatch} suggest that the output representations of single-view vision foundation models (VFMs) already exhibit strong emergent cross-image alignment capabilities, yet current pipelines still rely on additional matching modules to transform these representations into correspondences.
This observation, together with the stronger scalability of single-view pretraining, drive us to rethink: rather than initializing a ViT-based matcher from scratch through pair-centric pretraining, can it be bootstrapped from the far more scalable VFM itself?}

\begin{figure}[t]
  \centering
  \includegraphics[width=0.49\textwidth]{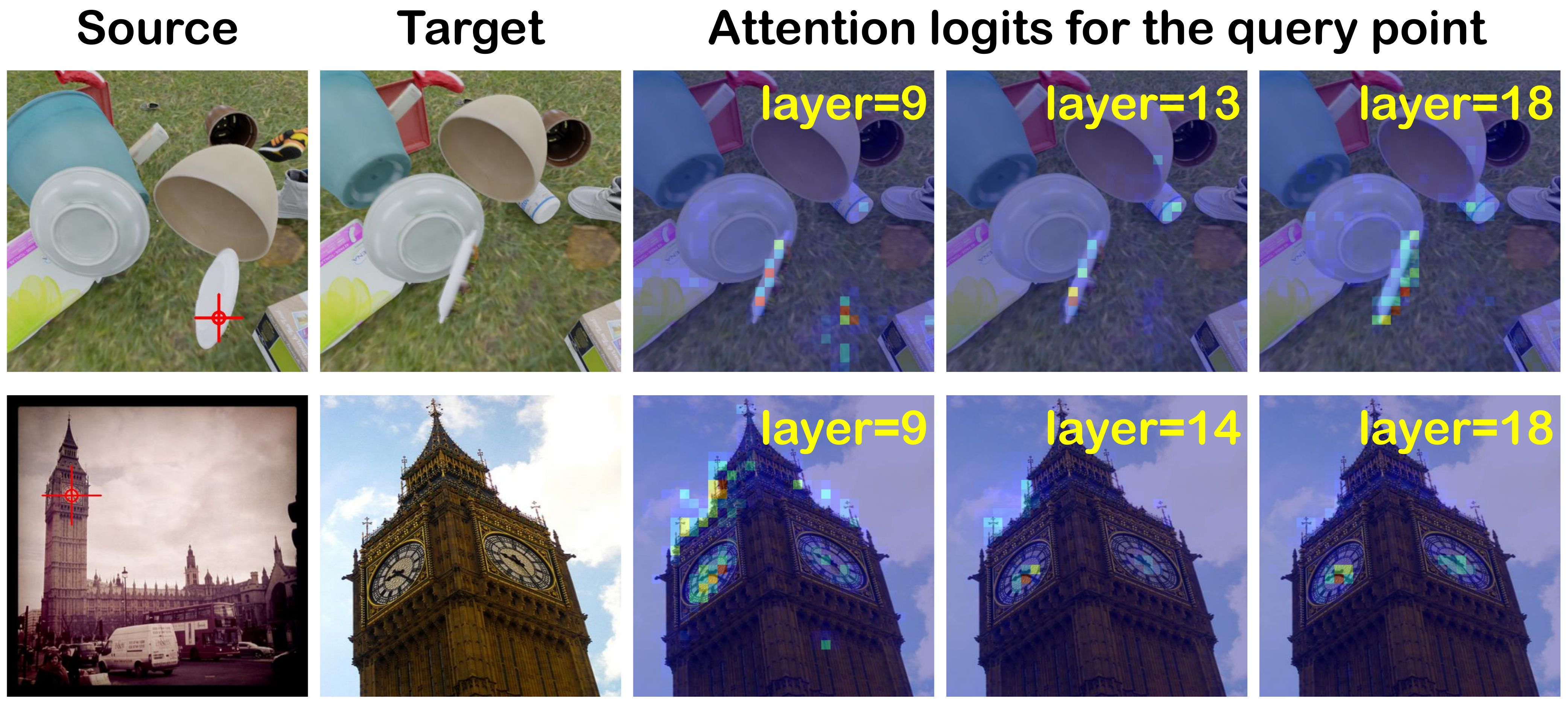}
  \caption{Zero-shot emergent matching capacity in DINOv3-Large~\cite{dinov3}. Cross-view attention logits (from source queries to target keys) are extracted at the layer-$th$ block. The emergent matching prior is robust to fast-moving objects (row 1) and significant appearance variations (row 2). Warmer colors denote higher attention affinities.
  }
  \label{fig:attn_vis}
\end{figure}

\textcolor{black}{
To explore this possibility, we investigate whether single-view VFMs internally encode transferable matching priors for matcher construction.}
Specifically, we probe this possibility by applying the query-key projections of a frozen VFM across two views. 
As illustrated in Fig.~\ref{fig:attn_vis}, the resulting attention logits concentrate on semantically corresponding regions, which inspires our first key \emph{insight: the query-key projections inside single-view VFMs encode coarse but reusable cross-view matching priors.}
% \textcolor{blue}{Going beyond previous findings~\cite{FormerStereo,RoMa} that VFMs output matchable representations across images, this observation reveals the matching prior within the attention projections.}
% By repurposing such pretrained single-view ViTs as the initialization for cross-view matchers, we bypass the need to learn cross-view representations from scratch and seamlessly inherit the scaling properties of single-view foundation models.

% \begin{figure}[t]
%   \centering
%   \includegraphics[width=0.5\textwidth]{./figures/pipeline_comparison.pdf}
%   \caption{Dense matching pipelines.}
%   \label{fig:pipeline_comparison}
% \end{figure}

\textcolor{black}{Given this finding, the next question is how to unleash the latent matching prior when adapting a pretrained single-view VFM into a matcher. % while preserving the pretrained ViT structure.
Since this prior is encoded in attention projections, a natural adaptation strategy is to reuse these projections for cross-view interaction, which can be implemented with (1) explicit cross-attention between views, or (2) global self-attention over concatenated views. 
Although both enable two-view interaction, they lead to drastically different behaviors at initialization. 
Explicit cross-attention directs each query to the opposite view and natively establish coarse cross-view affinity before fine-tuning.
In contrast, global self-attention jointly normalizes intra- and inter-view tokens, causing self-affinity to dominate the attention mass. 
This leads to our second \emph{insight: reusing pretrained projections in cross-attention form provides a structurally effective initialization for a ViT-based matcher built from a single-view VFM.}}
Inspired by these insights,
we present \textbf{F}low \textbf{A}ny \textbf{S}cene \textbf{T}ransformer (FAST), a scaling-oriented correspondence model. % built upon a \emph{vanilla} ViT.
% Instead of introducing a bespoke binocular pretraining architecture, 
\textcolor{black}{Rather than relying on bespoke matching modules or binocular pretraining,
FAST is born on a standard single-view VFM and is adapted to dense matching through a zero-parameter rewiring strategy, which reformulates selected self-attention layers into cross-attention ones.
To complement backbone scaling with data-side scaling, we assemble a 6-million-pair corpus, a similar scale to Croco but with additional optical-flow annotations, % through a depth-to-flow conversion procedure, 
therefore enabling task-specific adaptation after VFM-based initialization.}
Powered by these scalable design choices, FAST achieves state-of-the-art performance across diverse benchmarks.
More importantly, FAST scales favorably with both backbone size and training data, yielding consistent gains without the early saturation observed in volume-centric approaches~\cite{DCANet,SEA-RAFT}.
%These properties make FAST as a practical step toward a scalable correspondence backbone that can continuously benefit from single-view ViT foundation models and growing amounts of supervision.

Our contributions can be summarized as follows:
\begin{itemize}
\item \textcolor{black}{We reveal that the query-key projections inside single-view VFMs can be reused to produce coarse cross-view correlations, establishing the key basis for extending standard VFMs to dense matchers.}
\item \textcolor{black}{We introduce a zero-parameter attention rewiring strategy, which initializes a scalable correspondence matcher from readily available single-view VFMs.}
\item \textcolor{black}{We assemble Flow-6M, a 6-million-pair optical-flow dataset that consolidates existing flow data and converts static-scene geometry into dense flow labels.}%, enabling scalable supervision for FAST.}
\item We present FAST, a scalable correspondence foundation model that achieves strong generalization across \textcolor{black}{small- and wide-baseline matching benchmarks.}
\end{itemize}

\section{Related Work}
\label{sec:related_work}

% We categorize prior work into correlation-based and correlation-free paradigms by whether an explicit correlation/cost volume is constructed. In addition, we summarize a recent trend that leverages vision foundation models as robust representations or priors to enhance dense matching.

\subsection{Correlation-Based Method}
Correlation-based methods leverage correlation volumes to simplify the matching problem into a regression-from-classification task.
This paradigm has been widely adopted in both optical flow and stereo matching.

\noindent \textbf{Optical Flow.}
Dosovitskiy \textit{et al.}~\cite{FlowNet} first introduce a correlation layer to compute feature similarities for motion regression. 
Building on this volume-based design, subsequent studies devote substantial effort to developing correlation aggregation strategy to recover an accurate displacement distribution from the correlation volume.
Representative directions include coarse-to-fine estimation with multi-scale correlation volumes~\cite{PWCNet}, iterative motion refinement via repeated lookups in correlation volumes~\cite{RAFT}, and Transformer-integrated solutions~\cite{GMFlow,FlowFormer}.

\noindent \textbf{Stereo Matching.}
Stereo matching can be cast as a constrained correspondence problem where matches lie on epipolar lines. 
Early learning-based methods construct a 3D cost volume of size $D \times H \times W$ by evaluating matching costs over discrete disparity hypotheses~\cite{SceneFlow}. 
This strategy is widely adopted in efficient stereo algorithms~\cite{AANet,RAFT-Stereo} due to its memory-friendly design. 
Rather than compressing features into a scalar similarity, Kendall \textit{et al.}~\cite{GC-Net} construct a 4D volume of size $C \times D \times H \times W$ by concatenating cross-view features along all disparity proposals, followed by 3D CNNs for cost aggregation.
Subsequent works further enrich the volume formulation and aggregation through designs such as group-wise correlation, multi-scale volumes, and attention mechanisms~\cite{GwcNet,PSMNet,ACVNet}. 

% Overall, although correlation volumes encode geometric and correspondence priors, their construction and processing become prohibitively expensive as input resolution increases. This cost limits the capacity and complexity of cost aggregation modules and ultimately constrains matching performance.
\textcolor{black}{Overall, although correlation volumes provide explicit correspondence hypotheses, their volume-centric formulation tightly couples matching performance with task-specific aggregation and refinement modules. Simply stacking these modules often leads to diminishing returns, limiting the scalability of correspondence models beyond architecture-specific designs.}

\subsection{Correlation-Free Method}
% 实现correlation-free最直接的方法是将两张cross-view images沿着特征通道拼接，并直接回归motion. 然而，由于缺乏明确的xx，方法性能不好。为了明确实现两视图之间的信息交换，STTR怎么做；Croco怎么做。这些方法随着网络能力的增强而增强，但是网络缺乏明确的几何解释。我们的方法建立在可观测的几何线索上，因此效果更好，可解释性更强。
A straightforward correlation-free matching paradigm is to stack both input images together and process them through a network~\cite{FlowNet}. 
This design allows the network to decide how to fuse the image pair for motion regression. 
However, without an explicit cross-view interaction mechanism, such approaches typically struggle to achieve accurate motion estimation.
More recently, attention mechanisms have enabled new formulations for dense matching.
Along this line, 
Li \textit{et al.}~\cite{STTR} replace the cost volume with alternating self-attention and cross-attention along epipolar lines to aggregate cross-view evidence.
Weinzaepfel \textit{et al.}~\cite{croco,crocov2} propose a ViT-based encoder-decoder for two-view matching, coupled with cross-view masking pretraining and dense-matching-specific fine-tuning to obtain a correlation-free model.
Building on these advances, 
\textcolor{black}{
Leroy \textit{et al.}~\cite{MASt3R} introduce a local feature head to establish sparse correspondences for image matching.
Zhang \textit{et al.}~\cite{UFM} update the single-view encoder with DINOv2~\cite{dinov2} and establish cross-view interaction with additional global self-attention modules.
Edstedt \textit{et al.}~\cite{RoMav2} further improve the dense matching performance with frozen DINOv3, global self-attention, and cascaded warped refiners.
In parallel, recent advances~\cite{dust3r, monst3r, VGGT, depthanything3} demonstrate the effectiveness of attention for correlation-free 3D reconstruction, but they are primarily optimized for scene-level geometry rather than precise correspondence estimation.}
Following this attention-based direction, we realize cross-view interaction via attention. 
In contrast to prior work that designs train-from-scratch modules for dense matching, we directly leverage an existing pretrained vanilla ViT as the matcher.

\subsection{Scaling Route for Correspondence Models}
\noindent \textbf{Model Scaling with Foundation Backbones}.
\textcolor{black}{
Integrating vision foundation models (VFMs) into correspondence pipelines provides a natural route for model scaling.
Along this direction, recent advances either freeze the pretrained backbones as robust encoders~\cite{FormerStereo,PanMatch,RoMa}, adapt it with parameter-efficient modules~\cite{SMoE}, attach side-tuning branches~\cite{FoundationStereo}, or fully finetune the backbone~\cite{UFM} for downstream matching tasks.
Despite these different adaptation strategies, most methods follow an encoder-centric scaling paradigm, where VFMs are employed as single-view encoders, so that additional matching modules are still required for multi-view interaction.
As a result, increasing the VFM size mainly scales the feature extractor rather than the matching components. 
In contrast, we pursue a matcher-centric scaling route by repurposing a vanilla pretrained ViT as the matching network.}

\noindent \textbf{Data Scaling}.
\textcolor{black}{
Complementary to model scaling, data scaling has become another key route to improving correspondence generalization.
Weinzaepfel \textit{et al.}~\cite{croco} perform cross-view masked pretraining on millions of paired images to alleviate the data-hungry nature of ViT-based Croco model.
Wang \textit{et al.}~\cite{SEA-RAFT} demonstrate that motion pretraining on static scenes improves the generalization performance of optical flow estimation.
Recent efforts further enlarge the scale of labeled correspondence samples with conditional rendering~\cite{FoundationStereo}, depth-based warping~\cite{Flow-Anything}, multi-view geometric resampling~\cite{UFM}, and single-view homography transformations~\cite{MatchAnything}.
Building on this trend, we assemble a million-scale optical flow corpus, enabling our ViT-based matcher to benefit consistently from data scaling.}

\section{Method}
We present Flow Any Scene Transformer (FAST), a scalable dense correspondence model that transfers the scaling benefits of single-view foundation models to cross-view matching.
We first explain why a single-view pretrained VFM is effective for dense matching in Sec.~\ref{sec:rethinking}, then describe the rewired matching architecture in Sec.~\ref{sec:architecture}. 
The large-scale data corpus is introduced in Sec.~\ref{sec:data}, and the training objectives are presented in Sec.~\ref{sec:loss}.

\subsection{Rethinking Single-View ViTs for Dense Matching}
\label{sec:rethinking}
Recent advances~\cite{FormerStereo,SAMFlow,FoundationStereo} have demonstrated that single-view pretrained ViTs can act as robust feature extractors for dense matching.
However, within these pipelines, the matching process still relies on external correlation, followed by specialized cost aggregation and refinement modules to establish accurate correspondences.
Such a decoupled paradigm not only introduces substantial computational overhead, but also limits the network’s ability to scale in an end-to-end manner.
This limitation motivates a more fundamental question: instead of treating a pretrained ViT merely as a feature-extraction backbone, can the ViT itself serve as the matching network?

To answer this question, we first identify two core operations in dense matching: all-pairs correlation estimation across views, and matching evidence aggregation to resolve local ambiguities.
Fortunately, the computational substrate for both operations is inherent within the Transformer attention mechanism.
Given the query $\mathbf{Q}$ and key $\mathbf{K}$, the attention logits are computed as $\mathbf{M} = \mathbf{Q} \mathbf{K}^{\mathsf{T}} / \sqrt{D}$. Here, $\mathbf{Q}, \mathbf{K} \in \mathbb{R}^{N \times D}$ and $\mathbf{M} \in \mathbb{R}^{N \times N}$, where $N$ is the number of tokens and $D$ is the feature dimension. 
Once the token sequence is interpreted as a flattened image grid as $N = H \times W$, this inner-product formulation connects each query token to all key tokens via pairwise feature similarity, which is mathematically equivalent to constructing an all-pairs correlation volume over spatial patches.
Beyond correlation estimation, attention facilitates context-aware aggregation through value mixing, thereby integrating global cues in a manner that satisfies the requirements of matching evidence aggregation.

Given this structural equivalence, reusing self-attention kernels for cross-view matching only requires that corresponding tokens across views stay close in the shared feature space even under viewpoint and appearance changes.
This property is naturally encouraged by large-scale pretraining, which yields representations that are both semantically stable and spatially discriminative. 
Therefore, the pretrained query-key projections can be directly reused to measure cross-view token affinities without additional matching-specific initialization.

To empirically validate this point, we conduct a zero-shot probing experiment.
Specifically, given a cross-view image pair $\{\mathbf{x}, \mathbf{y}\}$, we pass both images through a frozen pretrained ViT, extract the projected query $\mathbf{Q}_{\mathbf{x}}$ and key $\mathbf{K}_{\mathbf{y}}$, and manually compute the cross-attention score map as
$
    \mathbf{M}_{\mathbf{x} \leftarrow \mathbf{y}}^{\text{cross}}=\mathbf{Q}_{\mathbf{x}} \mathbf{K}_{\mathbf{y}}^{\mathsf{T}} / \sqrt{D}
$.
As visualized in Fig.~\ref{fig:attn_vis},
strong attention responses frequently localize near the ground-truth corresponding regions, even though the model was never pretrained on explicit image-pair matching objectives. 
At the same time, these responses are still diffuse, noisy, and often multi-modal, which makes them insufficient for precise dense correspondence estimation without task-specific adaptation.
This discrepancy indicates that while the pretrained backbone inherently possesses a useful matching prior, it requires explicit cross-view interaction and task-specific fine-tuning to achieve accurate correspondences.

\begin{figure}[t]
  \centering
  \includegraphics[width=0.5\textwidth]{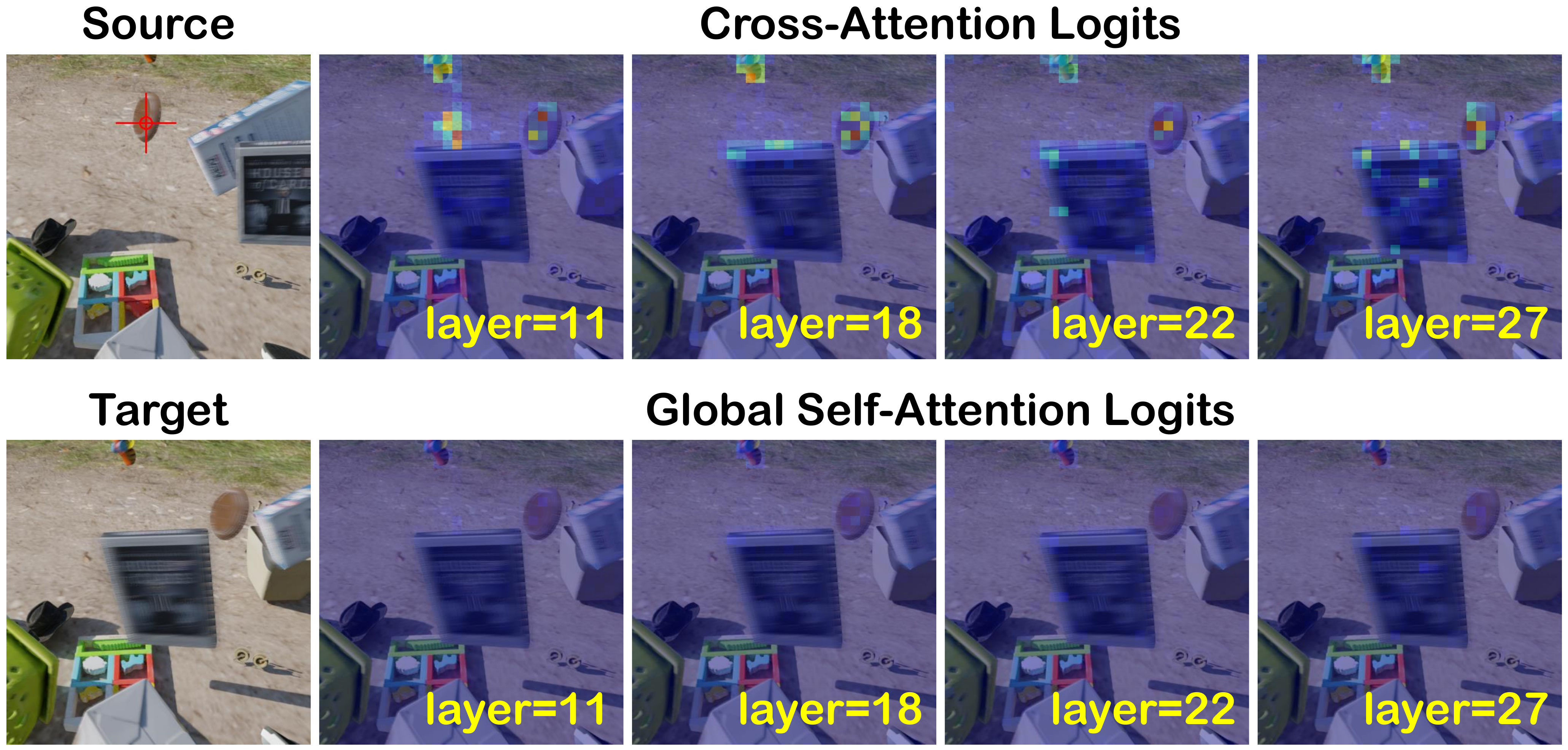}
  \caption{Comparison of attention responses at initialization. For a given query in the source view, we show its inter-view attention, including the cross-attention and global self-attention responses on the target view. All responses are directly generated using the Q and K projections from the layer-$th$ attention layer in DINOv3~\cite{dinov3}.
  }
  \label{fig:attention_comparison}
\end{figure}

\subsection{Employing ViTs as Matching Backbones}
\label{sec:architecture}
Building on the finding that single-view pretrained ViTs already encode a reusable matching prior,
the next critical step is to unleash this latent cross-view compatibility
without destroying the well-established inductive biases.
To enable cross-view interaction within a Transformer block, two straightforward paradigms can be considered:
(1) concatenating the two token sequences and applying global self-attention over the union of tokens, or
(2) keeping the two token sequences separate and introducing cross-attention across views.
The crucial difference between these paradigms lies in the normalization domain of the attention logits.
\textcolor{black}{We formalize this distinction below.}

\noindent \textbf{Analysis of Global Self-Attention.}
Let $\mathbf{x}, \mathbf{y} \in \mathbb{R}^{N \times D}$ denote the token sequences from the two views.
In global self-attention,
queries from view $\mathbf{x}$ attend to the concatenated token sequence $[\mathbf{x}; \mathbf{y}]$, yielding attention weights as
\begin{equation}
\mathbf{A}^{\mathrm{global}}_{\mathbf{x} \leftarrow \mathbf{xy}}
=
\operatorname{softmax}
\left(
\left[
  D^{-\frac{1}{2}}\mathbf{Q}_{\mathbf{x}} \mathbf{K}_{\mathbf{x}}^{\mathsf T};
  D^{-\frac{1}{2}}\mathbf{Q}_{\mathbf{x}} \mathbf{K}_{\mathbf{y}}^{\mathsf T}
\right]
\right),
\label{eq:global_attn_revised}
\end{equation}
where the softmax is applied jointly to both intra-view and cross-view tokens.
Since the backbone is pretrained for single-image perception, intra-view affinities tend to be dominant at initialization.
As a result, the cross-view logits are suppressed by the stronger intra-view responses within the shared softmax normalization, making the transferred model less likely to exploit cross-view evidence during the early stages of fine-tuning.

\noindent \textbf{Analysis of Cross-Attention.}
In contrast, explicit cross-attention restricts the normalization domain to the opposite view as
\begin{equation}
\mathbf{A}^{\mathrm{cross}}_{{\mathbf{x}} \leftarrow {\mathbf{y}}}
=
\operatorname{softmax}
\left(
D^{-\frac{1}{2}}\mathbf{Q}_{\mathbf{x}} \mathbf{K}_{\mathbf{y}}^{\mathsf T}
\right),
\label{eq:cross_attn_revised}
\end{equation}
thereby eliminating competition from intra-view tokens.
Guided by the inherent matching prior in QK projections,
this interaction establishes explicit all-pair correlations and
provides a more favorable initialization than global self-attention.
As illustrated in Fig.~\ref{fig:attention_comparison}, the initial cross-attention maps exhibit strong responses centered around the corresponding objects, whereas global self-attention yields negligible cross-view signals. These observations confirm that cross-attention is better aligned with the pretrained ViT for matching-oriented initialization.

\begin{figure}[t]
  \centering
  \includegraphics[width=0.4\textwidth]{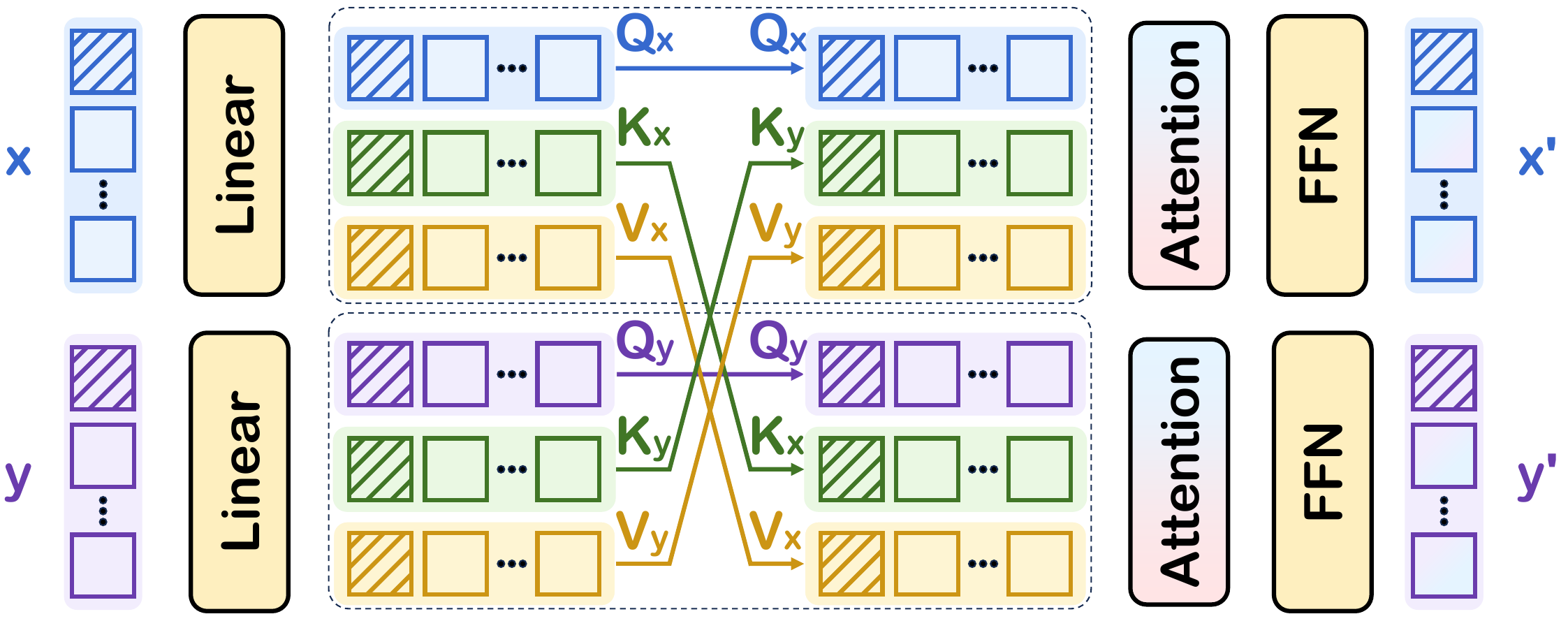}
  \caption{The proposed zero-parameter attention rewiring strategy, which converts self-attention to cross-attention.
  }
  \label{fig:attention_inference}
\end{figure}

\noindent \textbf{Zero-parameter attention rewiring.}
Motivated by the above analysis, we adopt cross-attention for cross-view interaction and implement it via a simple yet effective rewiring strategy.
For a standard self-attention layer, the output for view $\textcolor{blue}{\mathbf{x}}$ is given by
\begin{equation}
\operatorname{SA}(\textcolor{blue}{\mathbf{x}})
=
\operatorname{softmax}
\left(
D^{-\frac{1}{2}}
    \left( \textcolor{blue}{\mathbf{x}} \mathbf{W}_{Q} \right) 
    \left( \textcolor{blue}{\mathbf{x}} \mathbf{W}_{K} \right)^{\mathsf T}
\right)
\left(
    \textcolor{blue}{\mathbf{x}} \mathbf{W}_{V}
\right)
,
\label{eq:self_attn_revised}
\end{equation}
where $\mathbf{W}_Q$, $\mathbf{W}_K$, and $\mathbf{W}_V$ denote the QKV projections.
To convert this layer into cross-attention, we inherit these pretrained projections while swapping the sources of keys and values to the opposite view as
\begin{equation}
\operatorname{CA}(\textcolor{blue}{\mathbf{x}} \leftarrow \textcolor{purple}{\mathbf{y}})
=
\operatorname{softmax}
\left(
D^{-\frac{1}{2}}
    \left( \textcolor{blue}{\mathbf{x}} \mathbf{W}_{Q} \right) 
    \left( \textcolor{purple}{\mathbf{y}} \mathbf{W}_{K} \right)^{\mathsf T}
\right)
\left(
    \textcolor{purple}{\mathbf{y}} \mathbf{W}_{V}
\right).
\label{eq:rewired_ca_x}
\end{equation}
The symmetrical operation is applied to $\operatorname{CA}(\textcolor{purple}{\mathbf{y}}\leftarrow\textcolor{blue}{\mathbf{x}})$. 
Operationally, we process tokens from both views in parallel and modify only the token routing within the attention mechanism, as shown in Fig.~\ref{fig:attention_inference}.
This rewiring strategy introduces zero new parameters and explicitly leverages the matching prior encoded in query-key compatibility function, yielding a favorable initialization for cross-view matching.

\begin{figure}[t]
  \centering
  \includegraphics[width=0.48\textwidth]{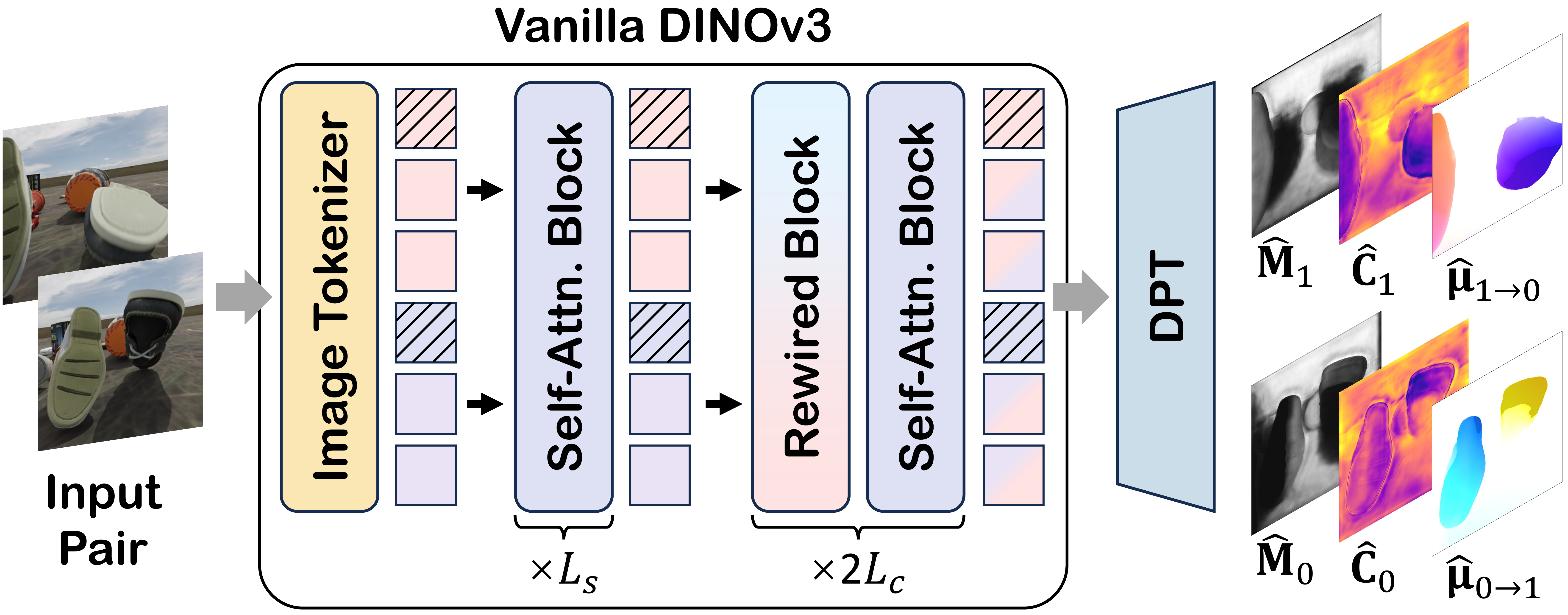}
  \caption{FAST consists only of a DINOv3~\cite{dinov3} backbone and a DPT head~\cite{DPT}. In a single forward pass, it jointly regresses bidirectional optical flow $\{\mathbf{\hat{\mu}}_{0 \rightarrow 1}, \mathbf{\hat{\mu}}_{1 \rightarrow 0}\}$, confidence maps $\{\mathbf{\hat{C}}_{0}, \mathbf{\hat{C}}_{1}\}$, and covisable mask $\{\mathbf{\hat{M}}_{0}, \mathbf{\hat{M}}_{1}\}$ from an RGB image pair.
  }
  \label{fig:architecture}
\end{figure}

\noindent \textbf{Overall Matching Architecture.}
Building on the rewiring strategy, we construct FAST on top of a plain single-view foundation model. 
We preserve standard self-attention in the first $L_s$ blocks to extract robust, viewpoint-invariant representations. The remaining blocks are organized into an alternating \texttt{CA--SA} pattern to formulate matcher.
The total block depth satisfies $L = L_s + 2L_c$, which exactly matches that of original ViT without additional Transformer blocks or parameters. 
Finally, the outputs from the $L_s$-$th$ and the last layers are rearranged as multi-level representations and fed into a DPT head~\cite{DPT} to progressively decode the dense correspondence field. 
The overall architecture is illustrated in Fig.~\ref{fig:architecture}.

\subsection{Large-Scale Supervision from Static Rigid Scenes}
\label{sec:data}
\textcolor{black}{
% Beyond scaling model capacity with larger pretrained backbones,
Beyond model scaling,
FAST also pursues data-side scaling by increasing both the number and the diversity of correspondence pairs.
Existing optical-flow datasets provide massive annotated image pairs, yet most samples are organized as video clips, where adjacent frames share repeated objects and similar scene layouts.
As a result, the large scale in pair count does not imply equally large scene-level and object-level diversity.
This motivates us to complement standard optical-flow data with additional co-visible image pairs from related geometric tasks.}

\textcolor{black}{
In practice, we consolidate heterogeneous correspondence sources, including rectified stereo and SfM/SLAM reconstructions, into a unified dense 2D displacement representation.
For stereo pairs, we convert disparity into horizontal or vertical displacement by applying random 90-degree rotations.
For calibrated static rigid scenes, we employ standard multi-view geometry to induce dense correspondences from depth and camera poses.
We further apply a depth-based consistency check to improve label reliability, retaining pixels that are projected inside the target image with consistent depth geometry.
The final corpus combines optical-flow data with converted static-scene data, expanding the scale and diversity of training pairs across diverse motions, scenes, and camera configurations.
% 后面的描述可去掉
% Optical-flow datasets are essential for learning dynamic motion patterns, including independently moving objects and non-rigid deformation. 
% On the other hand, the converted static-scene data expands the diversity of scenes, object categories, and camera baselines.
% The benefit of this diversity-oriented supervision design is validated in Sec.~\ref{sec:ablation}.
}

% \begin{figure}[t]
%   \centering
%   \includegraphics[width=0.45\textwidth]{./figures/depth_to_flow.pdf}
%   \caption{With the known depth and camera parameters, we can obtain pixel-wise correspondence across views through back-projection and projection.
%   }
%   \label{fig:depth_to_flow}
% \end{figure}

\subsection{Loss Function}
\label{sec:loss}
Following the probabilistic regression formulations~\cite{SEA-RAFT,crocov2}, we parameterize the optical-flow prediction as a Laplace distribution, where the estimated mean $\hat{\mu}$ and scale parameter $\hat{s}$ represent the expected flow and its uncertainty, respectively.
With this assumption, the network is optimized by minimizing the negative log-likelihood of the Laplace model as 
\begin{equation}
  \mathcal{L}_{Laplace}(\hat{\mu}, \hat{s})
  = - \log{\left[\frac{1}{2\hat{s}}\exp\left(\frac{- \vert \mu - \hat{\mu} \vert}{\hat{s}}\right)\right]},
  \label{eq:laplace}
\end{equation}
where $\mu$ denotes the optical-flow annotation. 
In practice, we predict the inverse scale as confidence, \textit{i.e.}, $\hat{c} = (\hat{s})^{-1}$, and estimate it in $\exp$ space to ensure the equivalent positive confidence $\hat{c} \in (0,+\infty)$. 
Assuming that $\mathbf{M}_{all}$ denotes all valid supervision containing a total of $N_{all}$ elements, Eq.~\ref{eq:laplace} can be rewritten as
\begin{equation}
  \mathcal{L}_{Laplace}(\hat{\mu}, \hat{c}) 
  = \frac{1}{N_{all}} \sum_{i \in \mathbf{M}_{all}} \left( \hat{c}_{i} \vert \mu_{i} - \hat{\mu}_{i} \vert - \log{\hat{c}_{i}} \right).
\end{equation}

\textcolor{black}{
Compared with a standard $\ell_1$ loss, the Laplace objective prevents loss domination by ambiguous regions, such as occlusions and noisy pseudo-labels~\cite{SEA-RAFT}.
However, this confidence-aware formulation may also allow the network to under-emphasize difficult yet matchable pixels by predicting low confidence.
This is undesirable for dense matching, where covisible pixels with large displacement or significant viewpoint changes should still be optimized.
To alleviate this issue, we introduce an additional confidence-independent robust regression loss~\cite{RoMa,UFM} on the covisible region $\mathbf{M}_{noc}$, which is formulated as
\begin{equation}
  \mathcal{L}_{robust}(\hat{\mu}) 
  = \frac{1}{N_{noc}} \sum_{i \in \mathbf{M}_{noc}} \mathcal{R}(\left\Vert \mu_{i} - \hat{\mu}_{i} \right\Vert_{2}).
\end{equation}
This term complements the Laplace loss by enforcing direct flow regression on pixels that are matchable, while keeping optimization stable for large residuals.
We additionally supervise the predicted covisibility mask $\hat{m}$ with a binary cross-entropy loss as
\begin{equation}
  \mathcal{L}_{noc}(\hat{m}) 
  = \frac{1}{N_{all}} \sum_{i \in \mathbf{M}_{all}} [- m_{i} \log\hat{m}_{i} + (1 - m_{i}) \log(1 - \hat{m}_{i})]
\end{equation}
where $m_{i}=1$ indicates that pixel $i$ is covisible, and $m_{i}=0$ otherwise. 
The covisable mask $\mathbf{M}_{noc}$ is obtained from forward-backward flow consistency check~\cite{UnFlow}.
The total training loss is formulated as 
$\mathcal{L} = \mathcal{L}_{Laplace} + \mathcal{L}_{robust} + \mathcal{L}_{noc}$.}

\section{Experiment}
\label{sec:experiment}

\subsection{Implementation}
\label{sec:implementation}

\noindent \textbf{Training Datasets.}
We collect a large amount of optical flow data  %~\cite{SceneFlow,Spring,VirtualKITTI2,Kubric,cvo,dynamicreplica,InfiniGen,sintel,FlowNet,AutoFlow} 
to facilitate learning diverse motion patterns. Besides, we convert disparity of synchronously captured stereo pairs %~\cite{crestereo,FoundationStereo,FallingThings,WMGStereo,KenBurns} 
or depth of static rigid scenes %~\cite{TartanAir,TartanAirv2,MatrixCity,GTA5-SfM,EDEN,Structured3D,replica,hypersim} 
into flow format to further enrich the object and scene diversity. The collected pairs with moving objects are over 0.6 million and the total training pairs are up to 6 million (see Appendix~E for more details).
\textcolor{black}{All training samples are grouped into three categories, including dynamic scenes, small-baseline static scenes and wide-baseline static scenes. During training, we sample uniformly across these categories.} 

\noindent \textbf{Training Strategies.}
By default, we employ DINOv3~\cite{dinov3} as the backbone.
% We train the model with mixed datasets, initially balancing datasets by sampling each dataset equally, then shift toward sampling in proportion to dataset size.
We set the total batch size equal to 16, and crop input pairs into 512$\times$512.
Models are trained with AdamW optimizer for 1000K iterations, with a one-cycle learning-rate schedule and maximum learning rate equal to $1 \times 10^{-4}$. 
During training, we adopt standard data augmentations~\cite{RAFT} to increase sample diversity.  %, including color jitter and image flipping to increase sample diversity. 
Training FAST-Huge takes 8 NPUs for 10 days.

\subsection{Benchmark Comparison on Optical Flow Task}
\label{sec:exp_flow}
We evaluate FAST-Huge on Sintel~\cite{sintel}, Spring~\cite{Spring}, and KITTI~\cite{kitti2015} benchmarks. Different from previous methods that often rely on benchmark-specific fine-tuning or dedicated checkpoints optimized for individual datasets, FAST-Huge is evaluated with a single unified checkpoint across all benchmarks without any dataset-specific adaptation.
As shown in Table~\ref{tab:optical_flow_benchmarks}, FAST-Huge demonstrates remarkable robustness across synthetic benchmarks. On the Sintel clean split, it achieves a state-of-the-art 0.94 EPE, improving over the previous best result by 12.1\%. On Spring, FAST-Huge delivers the lowest EPE, 1PE and F1 scores among all compared methods. 
Notably, FAST-Huge outperforms CrocoFlow across all metrics on Sintel and Spring, demonstrating the effectiveness of our scalable backbone and large-scale correspondence supervision.

\begin{figure}[t]
  \centering
  \includegraphics[width=0.45\textwidth]{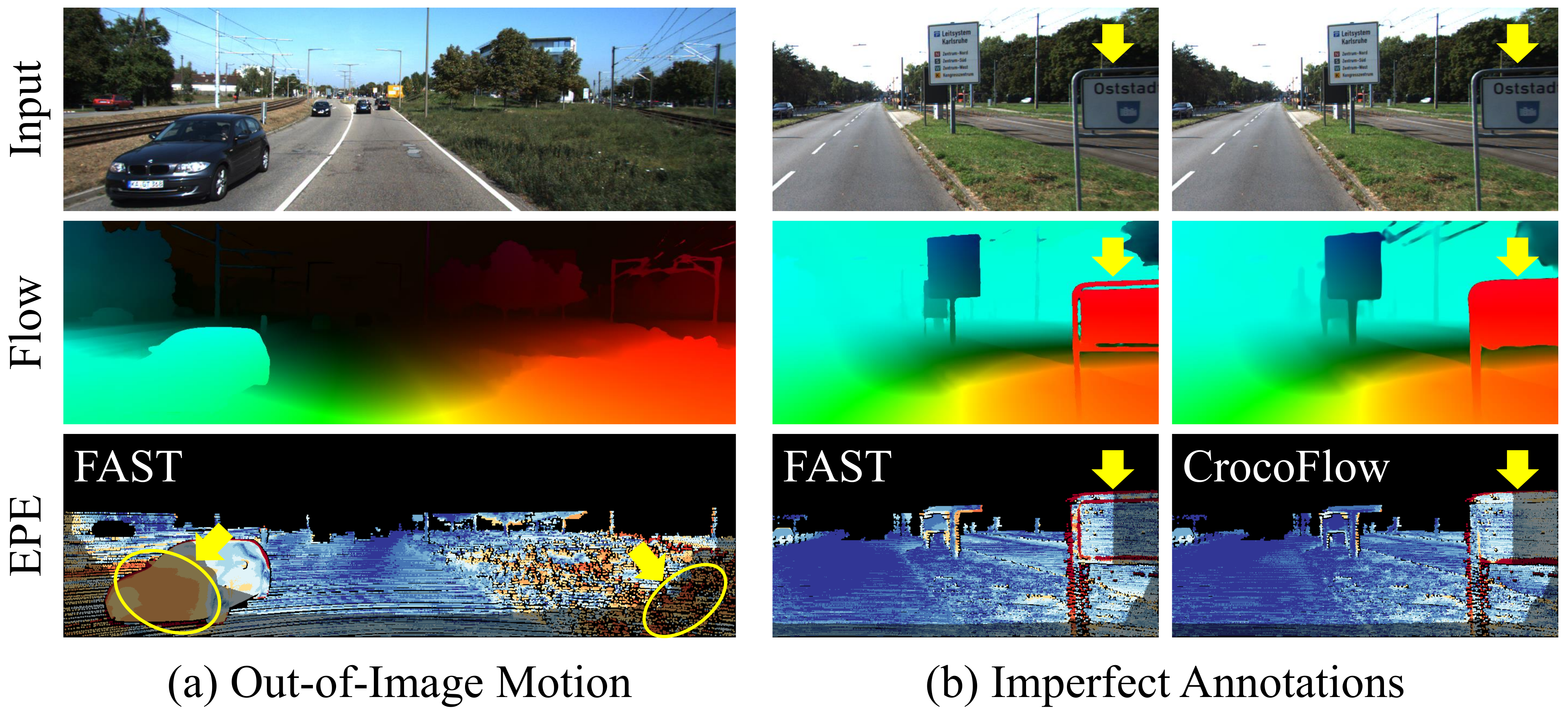}
  \caption{Analysis of the KITTI performance gap. FAST struggles with large out-of-image foreground motions, while its sharper boundary predictions are not fully reflected by KITTI metrics due to annotation ambiguity around object boundaries.
  }
  \label{fig:failure_case}
\end{figure}

On KITTI, FAST-Huge achieves competitive F1-background performance under zero-shot evaluation.
The remaining performance gap mainly originates from foreground motion estimation, where FAST is less effective for large out-of-image motions (Fig.~\ref{fig:failure_case}). 
Meanwhile, FAST produces sharper and better aligned motion boundaries, while such improvements are not fully reflected by KITTI metrics due to imperfect annotations around object boundaries~\cite{S2M2}.
Further increasing the diversity of dynamic scenes in training data and introducing sim-to-real self-supervised objectives may help mitigate this gap. 
Overall, these results highlight that FAST-Huge generalizes well across diverse flow benchmarks without task-specific fine-tuning.

\begin{table}[t]
  \centering
  \caption{Quantitative comparison on optical flow benchmarks. All metrics lower are better.}
  \setlength{\tabcolsep}{2pt}   % 增加列间距
  \begin{tabular}{lcccccccc}
  \toprule
  \multirow{2}{*}{Method} & \multicolumn{2}{c}{Sintel} & \multicolumn{3}{c}{Spring} & \multicolumn{3}{c}{KITTI} \\
  \cmidrule(lr){2-3}\cmidrule(lr){4-6}\cmidrule(lr){7-9}
    & clean        & final       & EPE     & 1PE     & F1     & F1-all  & F1-bg  & F1-fg  \\
    \midrule
    PWCNet~\cite{PWCNet}                  & 3.45         & 4.60        & 2.288   & 82.27   & 4.889  & 7.72    & 7.69   & 7.88   \\
    RAFT~\cite{RAFT}                    & 1.61         & 2.86        & 1.476   & 6.790   & 3.198  & 5.10    & 4.74   & 6.87   \\
    SEA-RAFT (M)~\cite{SEA-RAFT}            & 1.44         & 2.86        & \underline{0.363}   & \underline{3.686}   & \underline{1.347}  & 4.64    & 4.47   & \textbf{5.49}   \\
    CrocoFlow~\cite{crocov2}               & 1.09         & 2.44        & 0.498   & 4.565   & 1.508  & \textbf{3.64}    & \textbf{3.18}   & \underline{5.94}   \\
    FlowFormer++~\cite{FlowFormer++}            & \underline{1.07}         & \textbf{1.94}        & -       & -       & -      & \underline{4.52}    & -      & -      \\
    \textbf{FAST-Huge (Ours)}                    & \textbf{0.94}         & \underline{2.30}        & \textbf{0.323}   & \textbf{3.494}   & \textbf{1.212}  & 5.43    & \underline{3.99}   & 12.66 \\
    \bottomrule
  \end{tabular}
  \label{tab:optical_flow_benchmarks}
\end{table}

\begin{table}[t]
  \centering
  \caption{Zero-shot evaluation on stereo matching datasets.}
  \begin{tabular}{lrrrr}
  \toprule
  \multirow{2}{*}{Methods} & KT15 & KT12 & ETH3D & Midd. \\
                           & D1-all & D1-all & 1PE-noc & 2PE-noc \\
  \midrule
  Selective-IGEV~\cite{Selective-Stereo} & 4.5 & 3.2 & 3.4 & 7.5 \\
  MatchAttention-B~\cite{MatchAttention} & 3.78 & - & 1.27 & 2.27 \\
  S2M2-XL~\cite{S2M2} & 2.97 & 4.05 & \textbf{0.42} & \textbf{1.0} \\
  FoundationStereo~\cite{FoundationStereo} & \textbf{2.8} & \textbf{2.3} & 0.5 & 1.1 \\
  \midrule
  FlowFormer++~\cite{FlowFormer++} & 6.12 & 4.50 & 4.63 & 10.99 \\
  CrocoFlow~\cite{crocov2} & 4.10 & 3.83 & 2.59 & 7.08 \\
  Flow-Anything~\cite{Flow-Anything} & 4.12 & 3.08 & 3.32 & 8.53 \\
  PanMatch~\cite{PanMatch} & 3.27 & 2.77 & 1.79 & \textbf{3.39} \\
  % eth3d的结果需要放大两倍后用(768-448, 7680-448)的overlap计算得到
  \textbf{FAST-Huge (Ours)} & \textbf{2.97} & \textbf{2.51} & \textbf{1.24} & 3.84 \\
  \bottomrule
  \end{tabular}
  \label{tab:stereo matching}
\end{table}

\subsection{Zero-shot Evaluation on Stereo Matching Task}
\label{sec:exp_stereo}
% kitti 2012/15, eth3d, middlebury, drivingstereo, booster
% \noindent \textbf{Benchmarks and Metrics.} 
We evaluate the \emph{zero-shot} stereo matching capability of FAST-Huge on four standard benchmarks: KITTI 2012 \& 2015~\cite{kitti2012,kitti2015}, ETH3D~\cite{eth3d}, and Middlebury~\cite{middlebury}. 
% Following common practice, we report the threshold-based metrics: D1-All for KITTI, 2PE-noc for Middlebury, and 1PE-noc for ETH3D.
% The comparison is organized into two settings: flow-to-stereo transfer and stereo-to-stereo generalization.
% \noindent \textbf{Flow-to-Stereo Transfer.}
\textcolor{black}{
Specifically, we compare FAST-Huge with optical-flow methods that are enhanced by large-scale pretraining or finetuning, including FlowFormer++~\cite{FlowFormer++} (video pretraining on YouTube-VOS~\cite{xu2018youtube}), CrocoFlow~\cite{crocov2} (masked pretraining on 5.4M cross-view pairs), Flow-Anything~\cite{Flow-Anything} (motion finetuning on FA-Flow with 6M images), and PanMatch~\cite{PanMatch} (motion finetuning on 1.8M pairs).
This comparison assesses the task-transfer capability of general correspondence models from optical flow estimation to stereo matching.
As listed in Table~\ref{tab:stereo matching}, FAST-Huge outperforms existing optical-flow baselines across benchmarks, with the error rate reducing 9\% on KITTI15, 9\% on KITTI12 and 31\% on ETH3D.}

\textcolor{black}{We additionally report stereo matching models trained on million-scale stereo pairs as reference. Note that these methods explicitly exploit stereo geometry through components such as epipolar-constrained matching and warping-based refinement. Despite equipping without these designs, FAST-Huge outperforms Selective-IGEV on all four benchmarks, and surpasses MatchAttention-B on KITTI 2015 and ETH3D. 
Notably, FAST-Huge approaches FoundationStereo on the two KITTI benchmarks, with the absolute D1-all gaps only 0.21 on KITTI 2012 and 0.17 on KITTI 2015. 
Although FAST still lags behind on ETH3D and Middlebury with the hard 1pe and 2pe metrics, its scaling-oriented trends observed in Section 4.5 suggest that broader supervision and stronger backbones may further narrow the residual gap.}

% \noindent \textbf{Stereo-to-Stereo Generalization.}
% \textcolor{blue}{
% We next compare FAST-Huge with stereo matching models trained on million-scale stereo pairs, including Selective-IGEV~\cite{Selective-Stereo}, FoundationStereo~\cite{FoundationStereo}, MatchStereo-B~\cite{MatchAttention} and S$^2$M$^2$~\cite{S2M2}.
% These competitors incorporate stereo-specific designs, such as epipolar-constrained cost volume, warping-based refinement and 1/4-resolution inference. 
% In contrast, FAST adopts a much simpler formulation by rewiring a ViT as the generic 2D displacement matcher. 
% Nevertheless, after stereo-oriented fine-tuning, FAST-LoRA outperforms several scalable stereo baselines under a comparable trainable-parameter budget. 
% FAST-Huge with full fine-tuning further improves accuracy, yielding performance comparable to current state-of-the-art stereo methods.
% Although FAST still lags behind the strongest stereo-specialized model, its scaling-oriented and generic design provides a clear path toward further gains through larger foundation backbones and broader 3D correspondence supervision.}

% \begin{figure}[t]
%   \centering
%   \includegraphics[width=0.95\textwidth]{./figures/geometry_comparison.pdf}
%   \caption{Comparison of metric geometry. Depth Anything 3~\cite{depthanything3} produces visually plausible but metrically distorted structures, while ours strictly aligns with the ground-truth metric geometric relationships.
%   }
%   \label{fig:geometry_comparison}
% \end{figure}

\subsection{Wide-Baseline Benchmark Comparison}
\textcolor{black}{
The stereo and optical-flow evaluations mainly focus on small-baseline scenarios, where viewpoint changes are relatively limited. 
To further verify whether FAST can serve as a general-purpose dense correspondence model, we additionally evaluate it under wide-baseline settings. 
Specifically, we consider two evaluation protocols. 
First, we directly measure dense matching accuracy on DTU~\cite{DTU} and TA-WB~\cite{UFM}, and compare FAST with representative general matching methods~\cite{PanMatch,UFM,RoMav2}. 
Second, we evaluate relative pose estimation on WxBS~\cite{WxBS}, ScanNet~\cite{ScanNet}, and MegaDepth~\cite{MegaDepth}, where reliable camera pose recovery requires geometrically consistent matches under significant viewpoint and appearance variations. 
Note that FAST's results are obtained with the same checkpoint, without dataset-specific fine-tuning.}

\noindent \textbf{Evaluation on Dense Matching.}
\textcolor{black}{
We first compare FAST with UFM, a related ViT-DPT model augmented with 12 Transformer blocks and a refiner for matching.
As shown in Table.~\ref{tab:dense matching}, FAST outperforms UFM on all eight metrics, reducing EPE from 5.55 to 4.43 on DTU and from 12.84 to 12.03 on TA-WB.
Compared with RoMav2 which employs cascaded refiners, FAST achieves the lowest EPE metric on DTU and trails by only 0.7 and 3.2 percentage points under the 5pe criterion on DTU and TA-WB, respectively.
Together with Table~\ref{tab:optical_flow_benchmarks} \& \ref{tab:stereo matching}, these appended evidence suggest that FAST handles dense 2D correspondence across both small- and wide-baseline settings.}

% \usepackage{multirow}
\begin{table}[t]
  \centering
  \caption{Dense matching performance. Non-occluded metrics are reported.}
  \setlength{\tabcolsep}{4pt}
  \begin{tabular}{lcccccccc}
  \toprule
  \multirow{2}{*}{Method} & \multicolumn{4}{c}{DTU} & \multicolumn{4}{c}{TA-WB}  \\
  \cmidrule(lr){2-5} \cmidrule(lr){6-9} 
                          & epe & 1pe & 2pe & 5pe & epe & 1pe & 2pe & 5pe  \\
  \midrule
  PanMatch~\cite{PanMatch} & 26.69 & 60.7 & 44.8 & 32.7 & 74.36 & 69.7 & 55.8 & 48.0  \\
  UFM~\cite{UFM} & 5.55 & 55.5 & 32.9 & 13.8 & 12.84 & 51.4 & 30.6 & 17.0  \\
  RoMav2~\cite{RoMav2} & \underline{4.81} & \textbf{38.8} & \textbf{20.5} & \textbf{9.7} & \textbf{10.73} & \textbf{38.9} & \textbf{18.6} & \textbf{11.5}  \\
  FAST & \textbf{4.43} & \underline{54.0} & \underline{28.7} & \underline{10.4} & \underline{12.03} & \underline{51.0} & \underline{28.6} & \underline{14.7}  \\
  \bottomrule
  \end{tabular}
  \label{tab:dense matching}
\end{table}

\noindent \textbf{Relative Pose Estimation.}
\textcolor{black}{
We further evaluate FAST on pose estimation benchmarks, where accurate pose recovery requires reliable correspondences together with effective visibility/confidence estimates for filtering ambiguous matches.
As reported in Table~\ref{tab:pose estimation}, FAST achieves a second-best rank on the challenge WxBS benchmark, surpassing RoMav2 by 19.1 points.
FAST also obtains the second-best AUC@5$^\circ$/10$^\circ$ metric on ScanNet.
On MegaDepth, it consistently outperforms UFM under the same zero-shot setting, improving AUC@5$^\circ$/10$^\circ$/20$^\circ$ from 41.5/57.9/72.4 to 48.5/65.7/79.1.
These results  provide a stringent task-level verification of the accuracy and geometric consistency of FAST's dense matches.}

\begin{table}[t]
  \centering
  \caption{Relative pose estimation results on WxBS~\cite{WxBS}, ScanNet~\cite{ScanNet,SuperGlue}, and MegaDepth~\cite{MegaDepth,LoFTR}. We evaluate WxBS with mAA metrics while ScanNet and Megadepth with AUC metrics.}
  \label{tab:pose estimation}
  \setlength{\tabcolsep}{4pt}
  \begin{tabular}{lccccccc}
  \toprule
  \multirow{2}{*}{Method} 
  & \multicolumn{1}{c}{WxBS} 
  & \multicolumn{3}{c}{ScanNet} 
  & \multicolumn{3}{c}{MegaDepth} \\
  \cmidrule(lr){2-2} \cmidrule(lr){3-5} \cmidrule(lr){6-8} 
  & @10px & @5$^{\circ}$ & @10$^{\circ}$ & @20$^{\circ}$ 
  & @5$^{\circ}$ & @10$^{\circ}$ & @20$^{\circ}$ \\
  \midrule
  LightGlue~\cite{LightGlue} & --   & 17.8 & 34.0 & 52.0 & 51.0 & 68.1 & 80.7 \\
  LoFTR~\cite{LoFTR}     & 55.4 & 22.1 & 40.8 & 57.6 & 52.8 & 69.2 & 81.2 \\
  DKM~\cite{DKM}       & 58.9 & 29.4 & 50.7 & 68.3 & 60.4 & 74.9 & 85.1 \\
  RoMa~\cite{RoMa}      & \textbf{80.1} & \underline{31.8} & 53.4 & 70.9 & \underline{62.6} & \underline{76.7} & \underline{86.3} \\
  UFM~\cite{UFM}       & 42.3 & 31.3 & 54.1 & \underline{72.0} & 41.5 & 57.9 & 72.4 \\
  RoMav2~\cite{RoMav2}    & 55.4 & \textbf{33.6} & \textbf{56.2} & \textbf{73.8} & \textbf{62.8} & \textbf{77.0} & \textbf{86.6} \\
  \textbf{FAST}      & \underline{74.5} & \underline{31.8} & \underline{54.2} & 71.8 & 48.5 & 65.7 & 79.1 \\
  \bottomrule
  \end{tabular}
\end{table}

\subsection{Ablation Study}
\label{sec:ablation}
We conduct comprehensive ablation studies to validate: (1) the benefits of single-view pretraining for dense matching; (2) the superiority of our rewired cross-attention over global self-attention; (3) the decoder role; (4) the effect of trainable parameter scale; and (5) the scaling behavior with respect to both model size and training data.
We use DINOv3-Large as the backbone and train it on a combination of the SceneFlow, TartanAir, and Virtual KITTI 2 datasets for 1,000K iterations with a batch size of 8. 
We evaluate the zero-shot performance on the training splits of Middlebury, ETH3D, Sintel, and KITTI-15 datasets. Results are summarized in Table~\ref{tab:ablation}.

\begin{table}[t]
  \centering
  \caption{Zero-shot evaluation for ablation study. We ablate 
  {\color[HTML]{E67E22} initialization},
  {\color[HTML]{2AA198} attention},
  {\color[HTML]{3B5BDB} decoder role},
  % {\color[HTML]{8E44AD} loss design},
  {\color[HTML]{C0392B} parameter scale},
  {\color[HTML]{27AE60} model scale},
  and 
  {\color[HTML]{A37C00} data scale}
  across multiple datasets.
  \underline{Underlined} values denote the baseline settings, and \textbf{bold} indicates the best performance. $^*$ denotes evaluation on the training set. For all metrics, lower is better.
  }
  
  \setlength{\tabcolsep}{3.5pt}   % 增加列间距
  \begin{tabular}{@{}lcccccccccccc@{}}
  \toprule
  \multirow{3}{*}{Experiment}
  & \multicolumn{4}{c}{\textbf{Stereo Matching}}
  & \multicolumn{4}{c}{\textbf{Optical Flow}} \\

  \cmidrule(lr){2-5}\cmidrule(lr){6-9}
  
  & \multicolumn{2}{c}{\textbf{Middlebury}}
  & \multicolumn{2}{c}{\textbf{ETH3D  }}
  & \multicolumn{2}{c}{\textbf{Sintel}}
  & \multicolumn{2}{c}{\textbf{KITTI 15}} \\
  
  % 这一行也别 2-13 一条线，改成每个小组各画一段
  \cmidrule(lr){2-3}\cmidrule(lr){4-5}%
  \cmidrule(lr){6-7}\cmidrule(lr){8-9}
  
  & EPE & 2PE
  & EPE & 1PE
  & clean & final
  & EPE & F1-all \\
  
  \midrule  % pretraining
  {\color[HTML]{E67E22} Scratch}                          & 5.94 & 43.47 & 0.79 & 16.70 & 3.02 & 4.11 & 4.84 & 21.52 \\
  {\color[HTML]{E67E22} DINOv2~\cite{dinov2}}             & 2.01 & 16.84 & 0.37 & 5.02 & 1.21 & 2.28 & 2.27 & 8.85 \\
  % {\color[HTML]{E67E22} CroCov2~\cite{crocov2}}       & 2.42 & 14.45 & 0.29 & 3.65 & 1.55 & 2.47 & 2.11 & 7.21  \\
  % {\color[HTML]{E67E22} DAv3~\cite{depthanything3}}       & 1.89 & 13.12 & 0.26 & 2.58 & 2.01 & 3.09 & 5.64 & 44.26 \\
  {\color[HTML]{E67E22} \underline{DINOv3}~\cite{dinov3}}   & \textbf{1.77} & \textbf{13.18} & \textbf{0.28} & \textbf{3.10} & \textbf{1.06} & \textbf{2.13} & \textbf{2.06} & \textbf{7.33} \\% kitti是在 dr=(512x512) 的大小下测试的, 与 crocov2 保持一致
  
  \midrule  % attention
  {\color[HTML]{2AA198} global \texttt{SA}}           & 2.38 & 17.52 & 0.31 & 4.32 & 1.42 & 2.44 & 2.74 & 11.52 \\
  {\color[HTML]{2AA198} \underline{rewired \texttt{CA}}}   & \textbf{1.77} & \textbf{13.18} & \textbf{0.28} & \textbf{3.10} & \textbf{1.06} & \textbf{2.13} & \textbf{2.15} & \textbf{7.73} \\

  \midrule  % regression head
  % {\color[HTML]{3B5BDB} Linear-E2E}           & 3.08 & 31.87 & 0.51 & 10.12 & 1.57 & 2.52 & 3.05 & 15.28 \\   % 实验重做, 只需要冻结FAST, 替换DPT为linear
  {\color[HTML]{3B5BDB} Linear-Swap}           & 2.65 & 19.22 & 0.37 & 4.44 & 1.48 & 2.39 & 2.31 & 9.13 \\   % 实验重做, 只需要冻结FAST, 替换DPT为linear
  {\color[HTML]{3B5BDB} \underline{DPT}}   & \textbf{1.77} & \textbf{13.18} & \textbf{0.28} & \textbf{3.10} & \textbf{1.06} & \textbf{2.13} & \textbf{2.15} & \textbf{7.73} \\
  
  % \midrule  % loss
  % {\color[HTML]{8E44AD} $\mathcal{L}_{flow}$($L1$)}  & 2.10 & 16.89 & 0.31 & 3.91 & \textbf{1.05} & 2.17 & \textbf{2.14} & 8.34 \\
  % {\color[HTML]{8E44AD} $\mathcal{L}_{flow}$($\nabla$)}  & 2.05 & 13.64 & 0.28 & \textbf{2.72} & 1.08 & 2.17 & 2.30 & 8.24 \\
  % {\color[HTML]{8E44AD} \underline{$+ {\mathcal{L}_{depth}}$}}     & \textbf{1.77} & \textbf{13.18} & \textbf{0.28} & 3.10 & 1.06 & \textbf{2.13} & 2.15 & \textbf{7.73} \\

  \midrule  % lora
  {\color[HTML]{C0392B} LoRA(r=16)}   & 2.29 & 17.16 & 0.39 & 4.42 & 1.21 & 2.26 & 2.07 & 8.29 \\
  {\color[HTML]{C0392B} LoRA(r=64)}   & 1.91 & 14.09 & 0.28 & 3.25 & 1.09 & 2.05 & 2.14 & 8.23 \\
  {\color[HTML]{C0392B} \underline{full}}         & \textbf{1.77} & \textbf{13.18} & \textbf{0.28} & \textbf{3.10} & \textbf{1.06} & \textbf{2.13} & \textbf{2.15} & \textbf{7.73} \\

  \midrule  % model scale
  {\color[HTML]{27AE60} Base}           & 2.14 & 18.32 & 0.35 & 4.63 & 1.28 & 2.34 & 2.90 & 11.26 \\
  {\color[HTML]{27AE60} \underline{Large}}   & 1.77 & 13.18 & 0.28 & 3.10 & 1.06 & 2.13 & 2.15 & 7.73 \\
  {\color[HTML]{27AE60} Huge}           & \textbf{1.36} & \textbf{10.99} & \textbf{0.26} & \textbf{2.65} & \textbf{0.98} & \textbf{2.01} & \textbf{1.90} & \textbf{7.34} \\

  \midrule  % data scale
  {\color[HTML]{A37C00} SF}                    & 2.62 & 18.60 & 0.37 & 5.52 & 1.05 & 2.82 & 6.24 & 29.76 \\
  {\color[HTML]{A37C00} \underline{SF+T+VK}}   & 1.77 & 13.18 & 0.28 & 3.10 & 1.06 & 2.13 & 2.15 & 7.73 \\
  {\color[HTML]{A37C00} Flow-Dynamic}   & 1.97 & 16.58 & 0.32 & 4.30 & 1.04$^*$ & 1.59$^*$ & 2.13 & 7.78 \\
  {\color[HTML]{A37C00} Flow-6M}   & \textbf{1.11} & \textbf{7.59} & \textbf{0.26} & \textbf{2.34} & \textbf{0.83}$^*$ & \textbf{1.39}$^*$ & \textbf{2.05} & \textbf{7.60} \\

  \midrule
  FAST-Huge                             & \cellcolor[HTML]{FFCE93}1.02 & \cellcolor[HTML]{FFCE93}6.41 & \cellcolor[HTML]{FFCE93}0.22 & \cellcolor[HTML]{FFCE93}2.06 & \cellcolor[HTML]{FFCE93}0.67$^*$ & \cellcolor[HTML]{FFCE93}1.12$^*$ & \cellcolor[HTML]{FFCE93}1.82 & \cellcolor[HTML]{FFCE93}6.13 \\
  \bottomrule
  \end{tabular}
  \label{tab:ablation}
\end{table}

\noindent \textbf{{\color[HTML]{E67E22} Scalable Pretraining Initialization.}}
Large-scale single-view pretraining is the cornerstone of FAST. As shown in Table~\ref{tab:ablation}, training the ViT backbone from scratch drastically degrades generalization, indicating that pretraining is essential for modeling precise correspondence.
Furthermore, upgrading the initialization from DINOv2-Large (pretrained on 142M images) to DINOv3-Large (pretrained on 1.6B images) yields consistent performance improvements across all metrics. 
These results demonstrate that FAST effectively inherits the scaling benefits of single-view pretraining, where larger-scale pretrained representations provide increasingly stronger initialization for dense matching.

\noindent \textbf{{\color[HTML]{2AA198} Cross-view Interaction.}}
We compare our rewired cross-attention against global self-attention for two-view interaction. 
Although global self-attention also enables effective cross-view interaction and achieves competitive matching performance, replacing it with rewired cross-attention consistently reduces estimation errors across all evaluation metrics. These results indicate that explicitly modeling cross-view affinities through cross-attention is better aligned with correspondence matching than jointly modeling intra- and inter-view interactions with global self-attention.

\begin{figure}[t]
  \centering
  \includegraphics[width=0.49\textwidth]{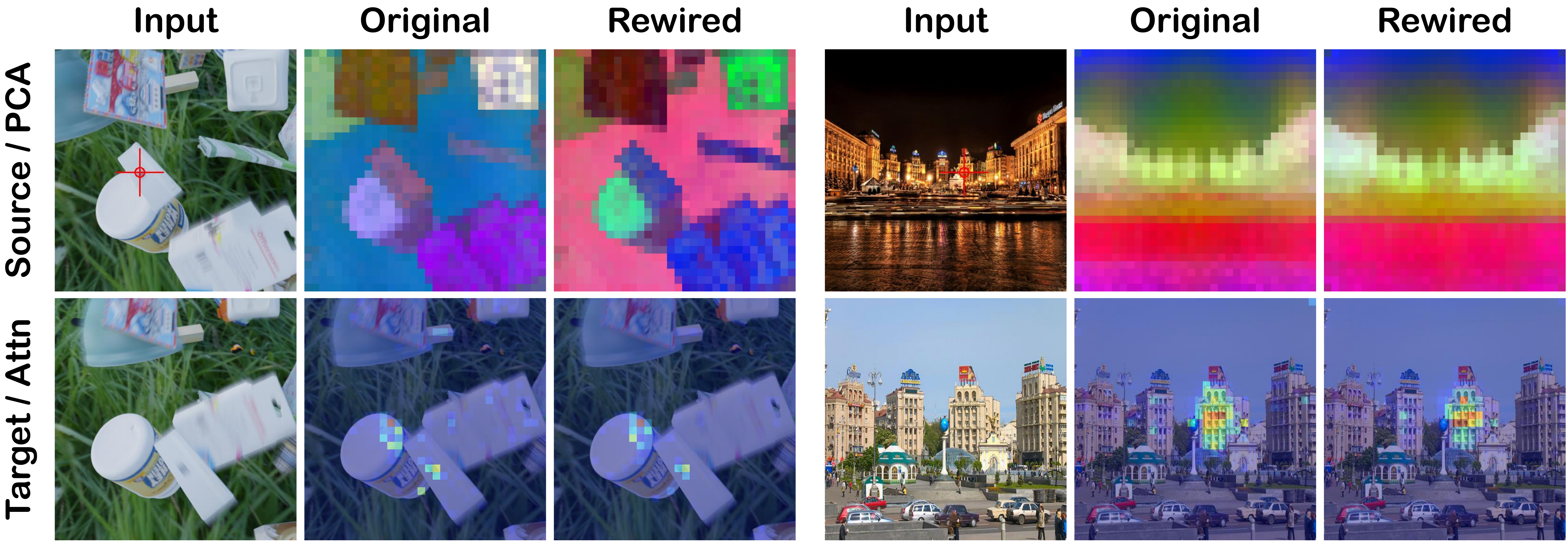}
  \caption{Last-layer comparison between original and rewired DINOv3-Huge (rewired from layer 15 of total layer 32).
  }
  \label{fig:rewired_init_impact}
\end{figure}

Furthermore, to examine whether the attention rewiring mechanism preserves the pretrained ViT's inductive bias at initialization, we visualize the last-layer representations and attention responses before and after rewiring. As shown in Fig.~\ref{fig:rewired_init_impact}, despite significant cross-view variations caused by object motion and appearance changes, the PCA structures and correspondence-related attention patterns remain consistent after converting self-attention into cross-attention. This observation suggests that the matching prior encoded in DINOv3 is preserved during the transition from single-view perception to correspondence matching.

\noindent \textbf{{\color[HTML]{3B5BDB} Decoder Role.}}
% \textcolor{blue}{
% We ablate the decoder to disentangle its contribution from the backbone's matching capability.
% Specifically, we construct two variants:
% (1) replacing DPT with a lightweight linear head and training FAST end-to-end; and
% (2) replacing DPT in a trained FAST-DPT model with an initialized linear head while keeping the rewired ViT frozen.
% As shown in Table~\ref{tab:ablation},
% the end-to-end linear variant achieves non-trivial zero-shot performance, indicating that FAST can establish effective correspondences without relying on a high-capacity decoder.
% The head-swap variant also retains meaningful matching accuracy, demonstrating that the learned cross-view matching capacity is encoded in the rewired ViT representations rather than memorized by the DPT head. 
% The remaining performance gap mainly comes from DPT's stronger multi-level decoding and sub-pixel refinement capabilities, as detailed in APPENDIX~E.}
\textcolor{black}{
To identify whether the learned correspondence is encoded in the rewired ViT or depends on the decoder capacity, we start from a trained FAST, replace its DPT (44M params.) with a linear head (1.4M params.), freeze the ViT backbone and train only the new head. As shown in Table~\ref{tab:ablation}, despite 31$\times$ compression in parameters, Linear-Swap variant increases EPE by only 0.09\text{--}0.88 pixels. This modest degradation indicates that correspondence is decodable from the frozen ViT representations rather than memorized by DPT. The remaining performance gap mainly comes from DPT's stronger multi-level decoding and sub-pixel refinement capabilities.}%, as detailed in Appendix~E.}

% \noindent \textbf{{\color[HTML]{8E44AD} Loss Design.}}
% We observe that a standard $L1$ loss $\mathcal{L}_{flow}(L1)$ already provides a strong zero-shot baseline.  Replacing it with Laplace regression objective, $\mathcal{L}_{flow}(\nabla)$, further mitigates estimation errors, notably reducing the Middlebury 2PE metric from 16.89\% to 13.64\%. Moreover, incorporating auxiliary depth supervision ($+ \mathcal{L}_{depth}$) yields additional improvement and achieves the best overall performance.
% These results demonstrate the efficacy of our combined loss design.

\noindent \textbf{{\color[HTML]{C0392B} Parameter-Efficient Adaptation.}}
\textcolor{black}{
We compare full fine-tuning with LoRA-based adaptation to investigate whether FAST relies on full backbone fine-tuning to acquire dense matching capability.
As shown in Table~\ref{tab:ablation}, LoRA with $r=16$ reduces the trainable ViT parameters from 384M to only 4M.
Nevertheless, LoRA adaptation still yields strong zero-shot performance, and further approaches the fully fine-tuned counterpart when increasing the rank from 16 to 64. This result suggests that the transferred dense matching ability mainly comes from the pretrained ViT representations and the rewired cross-attention formulation, rather than from unconstrained parameter updates.}

\noindent \textbf{{\color[HTML]{27AE60} Model Scalability.}}
To evaluate model scalability, we instantiate FAST with DINOv3 backbones of increasing sizes: Base, Large, and Huge. As shown in Table~\ref{tab:ablation}, performance improves across all zero-shot benchmarks as the model scale increases. Notably, we observe no performance saturation within the evaluated range, indicating that FAST effectively inherits the scaling properties of foundation models, translating larger pretrained ViTs into proportionally stronger dense matching capabilities.

\noindent \textbf{{\color[HTML]{A37C00} Data Scalability.}}
We explore data scalability by training FAST on progressively larger datasets:
FlyingThings3D (0.07M samples), 
a mixture of SceneFlow, TartanAir, and Virtual KITTI 2 (0.5M samples), 
Flow-Dynamic (0.6M dynamic-scene pairs), 
and Flow-6M (Flow-Dynamic augmented only with static-scene pairs). 
Results in Table 5 support two conclusions. 
First, performance improves overall as the training corpus scales from 0.07M to 6M pairs, confirming that FAST benefits effectively from data scaling. 
Second, augmenting Flow-Dynamic solely with static-scene pairs improves all reported metrics, including those on the dynamic Sintel and KITTI-15 benchmarks. 
This demonstrates that static-scene supervision strengthens geometric matching priors that transfer to independent object motion.

%--------------------------------------------------------
%
%--------------------------------------------------------
\section{Conclusion}
\label{sec:conclusion}
We reveal that single-view VFMs encode transferable cross-view matching priors, which can be effectively reused in cross-attention form through zero-parameter attention rewiring.
Building on this finding, we introduce FAST, a scaling-oriented correspondence model built from DINOv3. 
We further assemble Flow-6M to support data-side scaling.
Extensive experiments show that FAST achieves state-of-the-art performance across benchmarks, while scaling favorably with both backbone capacity and training data.

% or


\appendices
\section{Robust Regression Loss}
The robust regression loss is formulated as
\begin{equation}
  \mathcal{R}(x)
  = \frac{\left| \alpha - 2 \right|}{\alpha} \left( \left( \frac{\left( \frac{x}{\beta} \right)^{2}}{\left| \alpha - 2 \right|} + 1 \right)^{\frac{\alpha}{2}} - 1 \right),
\end{equation}
where we set $\alpha=0.5$ and $\beta=0.24$ following UFM~\cite{UFM}'s choice.

\section{Depth-to-Flow Label Generation}
\label{sec:depth-to-flow}
Given the depth map $\mathbf{D}_0$ and camera intrinsic matrix $\mathbf{K}_0$ of view 0, we first back-project each pixel $\mathbf{p}=(u_0,v_0)$ into 3D camera coordinates as
\begin{equation}
    \mathbf{P}_0 = \mathbf{D}_0(\mathbf{p}) \mathbf{K}_0^{-1} \tilde{\mathbf{p}}, \quad
    \tilde{\mathbf{p}} = [u_0, v_0, 1]^{\mathsf{T}},
\end{equation}
where $\mathbf{P}_{0}$ represents the 3D point corresponding to pixel $\mathbf{p}$ in the camera coordinate system of view 0.
Let the extrinsics of views 0 and 1 be $[\mathbf{R}_0 | \mathbf{t}_0]$ and $[\mathbf{R}_1 | \mathbf{t}_1]$, respectively, under the world-to-camera convention. The relative transformation $\mathbf{T}_{1 \leftarrow 0} = [\mathbf{R}_{1 \leftarrow 0} | \mathbf{t}_{1 \leftarrow 0}]$ from view 0 to view 1 is given by
\begin{equation}
    \mathbf{R}_{1 \leftarrow 0} = \mathbf{R}_{1} \mathbf{R}_{0}^{-1}, \quad
    \mathbf{t}_{1 \leftarrow 0} = \mathbf{t}_{1} - \mathbf{R}_{1 \leftarrow 0} \mathbf{t}_{0}.
\end{equation}
The 3D point $\mathbf{P}_{0}$ is subsequently transformed into the coordinate frame of view 1 and projected onto its image plane as
\begin{equation}
    s \tilde{\mathbf{p}}_{1} = \mathbf{K}_{1} (\mathbf{R}_{1 \leftarrow 0} \mathbf{P}_0 + \mathbf{t}_{1 \leftarrow 0}), \quad
    \tilde{\mathbf{p}}_{1} = [u_1, v_1, 1]^{\mathsf{T}},
\end{equation}
where $s$ denotes the derived depth of the transformed 3D point in the coordinate frame of view 1. 
This projection can be equivalently written in homogeneous form as $\tilde{\mathbf{p}}_{1} \sim \mathbf{K}_{1} \mathbf{T}_{1 \leftarrow 0} \mathbf{P}_{0}$.
With the calculated image coordinates $\mathbf{p}_1 = (u_1, v_1)$, the optical flow from view 0 to view 1 is thus defined as $\mathbf{F}_{1 \leftarrow 0} = \mathbf{p}_1 - \mathbf{p}$. The reverse flow $\mathbf{F}_{0 \leftarrow 1}$ is obtained analogously.
Finally, by comparing the derived depth $s$ with the ground-truth depth $\mathbf{D}_1(\mathbf{p}_1)$, we can identify occluded regions as well as geometrically inconsistent correspondences.
Specifically, if the projected depth $s$ is significantly greater than the observed depth $\mathbf{D}_1(\mathbf{p}_1)$, it implies that the 3D point is occluded by a foreground object in view 1, allowing us to accurately filter out non-covisible regions.

%---------------------------------------------------------------------------------
\section{Architecture Comparison}
We summarize the architectural compositions of representative ViT-based correspondence models in Table~\ref{tab:architecture_comparison}. Existing approaches typically employ a pretrained image encoder followed by an additional parameterized module for cross-view interaction. For instance, CroCo v2~\cite{croco, crocov2} adopts a Base scale ViT as the cross-attention decoder on top of a ViT-L encoder, while UFM~\cite{UFM} and RoMa v2~\cite{RoMav2} introduce additional Transformer blocks to exchange information between independently extracted foundation-model features. In contrast, FAST establishes cross-view interaction directly within the pretrained backbone by rewiring selected self-attention layers into cross-attention. This design reuses the pretrained attention projections as matching modules and therefore introduces zero additional parameters for cross-view interaction.

\begin{table}[t]
\centering
\caption{Architecture comparison of representative ViT-based 
correspondence models. ``Params.'' denotes parameters newly 
introduced for cross-view interaction, excluding the pretrained 
backbone and dense prediction heads.}
\label{tab:architecture_comparison}
\setlength{\tabcolsep}{2pt}
\begin{tabular}{lllrl}
\toprule
Method & Backbone & Cross-view Interaction & Params. & Decoder \\

\midrule
CroCo v2~\cite{crocov2} & ViT~\cite{ViT} & ViT-B & 114M & DPT \\
UFM~\cite{UFM} & DINOv2~\cite{dinov2} & 12$\times$ Transformer & 86M & DPT \\
RoMa v2~\cite{RoMav2} & DINOv3~\cite{dinov3} & 12$\times$ Transformer + CNN & 90M & DPT + Linear \\
FAST (Ours) & DINOv3~\cite{dinov3} & Rewired Transformer & \textbf{0} & DPT \\
\bottomrule
\end{tabular}
\end{table}

%-----------------------------------------------------------------------------
\section{Qualitative Comparison and Downstream Application}
In this section, we provide additional qualitative comparisons to complement the quantitative evaluations in the main paper. We consider increasingly challenging correspondence settings, ranging from optical flow and rectified stereo matching to wide-baseline image matching, and finally evaluate FAST as a correspondence front-end for multi-view 3D reconstruction. 
Unless otherwise specified, all FAST results are produced using the same unified FAST-Huge checkpoint without benchmark-specific fine-tuning or adaptation.

\subsection{Qualitative Comparison on Optical Flow}
We first compare FAST with CroCo-Flow~\cite{crocov2} and SEA-RAFT~\cite{SEA-RAFT} on the Sintel~\cite{sintel} and Spring~\cite{Spring} benchmarks. 
CroCo-Flow provides a particularly relevant comparison, as both methods follow a ViT-based dense prediction paradigm with a DPT head. 
However, CroCo-Flow relies on dedicated cross-view pretraining to initialize its binocular architecture, whereas FAST directly repurposes a pretrained single-view ViT as the matching backbone through attention rewiring, without requiring cross-view pretraining. 
We additionally include SEA-RAFT as a representative correlation-based optical-flow method, providing a complementary comparison with a specialized flow architecture.

\begin{figure}[t]
  \centering
  \includegraphics[width=0.45\textwidth]{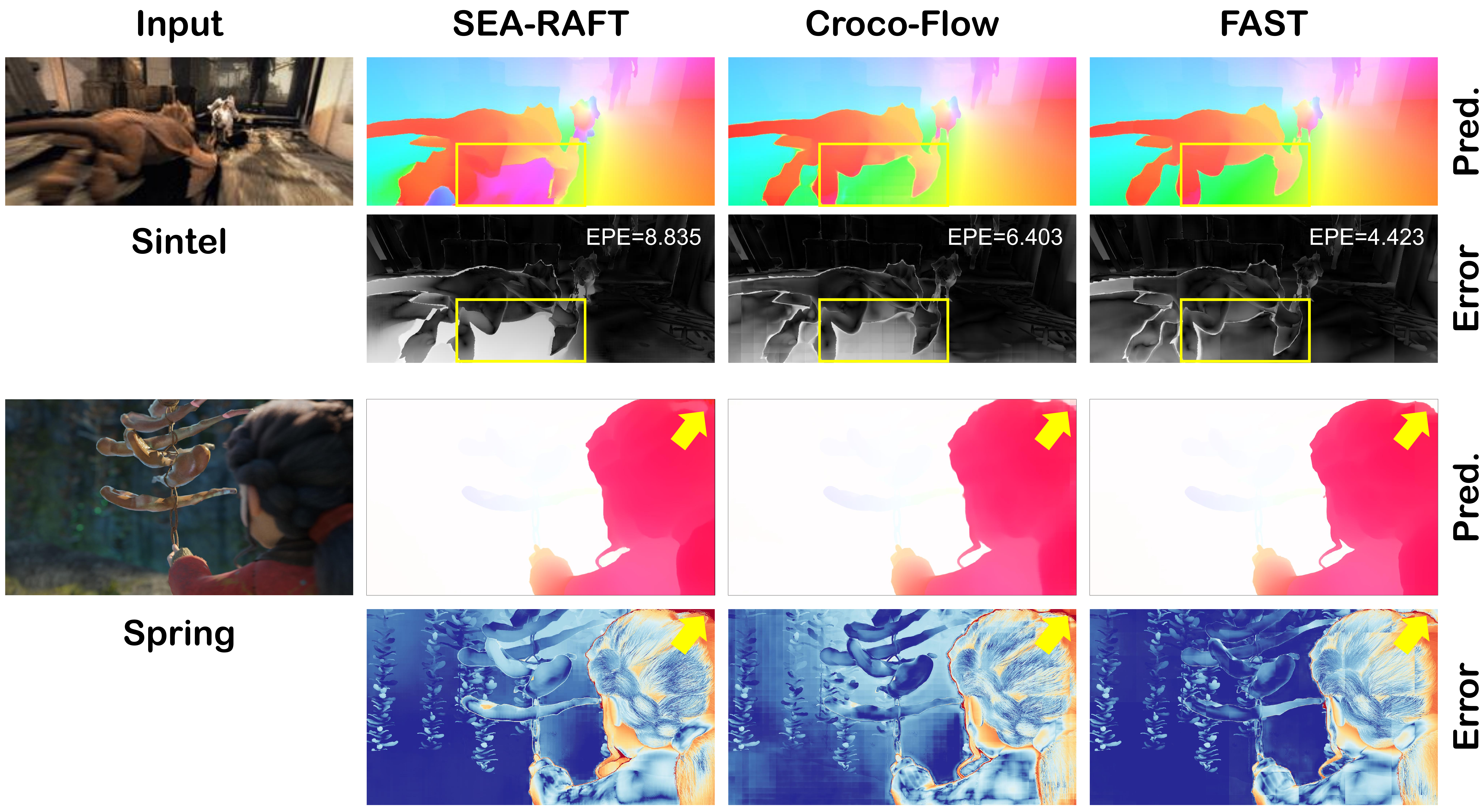}
  \caption{Optical flow estimation on Sintel and Spring scenes.
  }
  \label{fig:flow_comparison}
\end{figure}

As shown in Fig.~\ref{fig:flow_comparison}, FAST exhibits clear advantages in challenging motion regions. On Sintel, FAST better handles large occluded regions and produces a more spatially coherent flow field, reducing the overall EPE to 4.423, compared with 6.403 for CroCo-Flow and 8.835 for SEA-RAFT.
On Spring, FAST produces sharper and better image-aligned motion boundaries, particularly along the highlighted foreground contour, while exhibiting noticeably lower errors over the background region. These results suggest that FAST not only improves overall matching accuracy, but also better preserves motion discontinuities in challenging high-resolution scenes.

\subsection{Zero-shot Generalization on Stereo Matching}
\begin{figure}[t]
  \centering
  \includegraphics[width=0.45\textwidth]{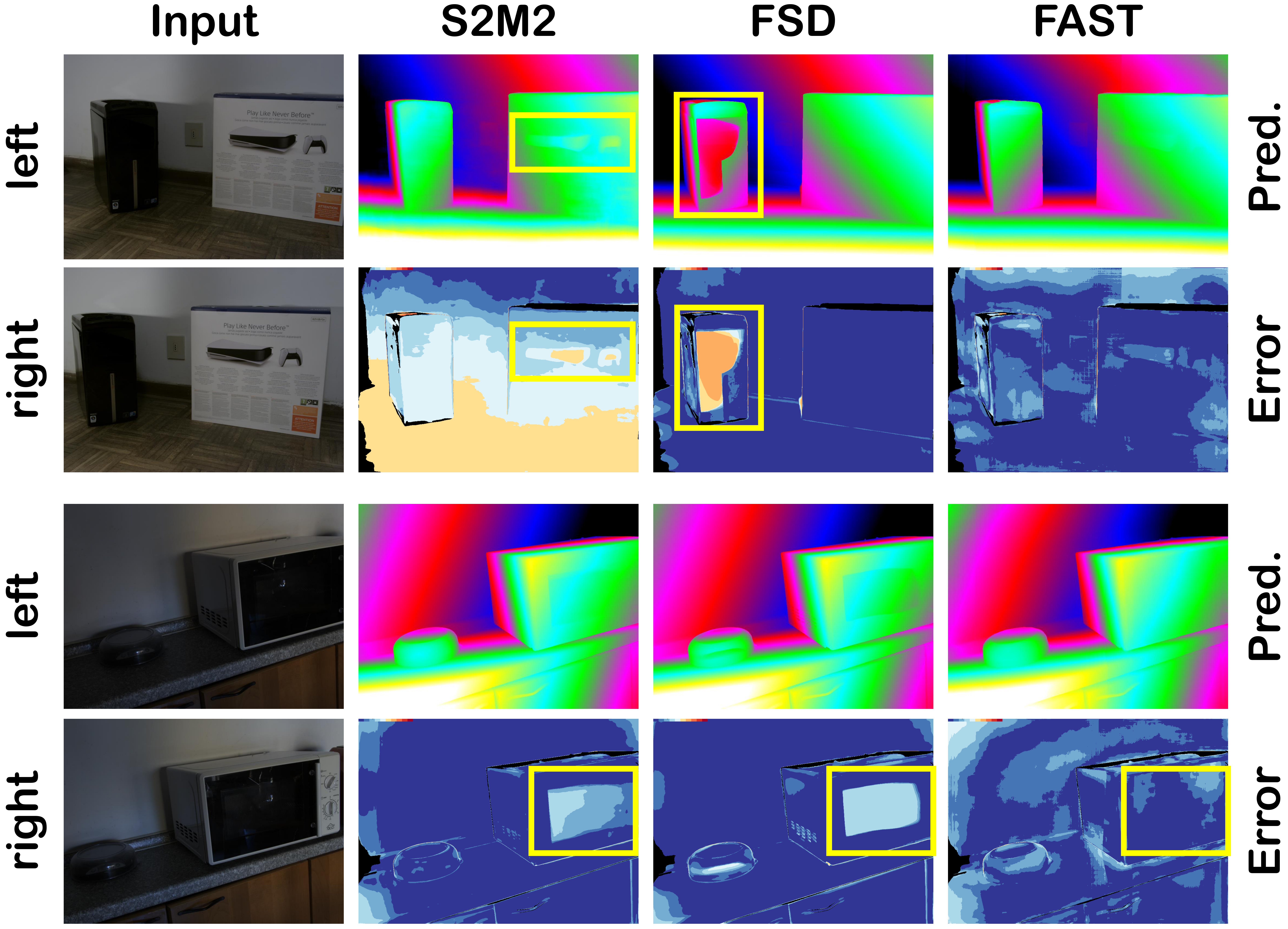}
  \caption{Stereo matching estimation on Booster scenes.
  }
  \label{fig:stereo_comparison}
\end{figure}

We next compare FAST with S$^2$M$^2$~\cite{S2M2} and FoundationStereo~\cite{FoundationStereo}, two recent state-of-the-art stereo matching methods. 
Unlike these stereo-specialized models, FAST directly transfers its generic 2D correspondence capability to rectified stereo matching without relying on stereo-specific cost-volume reasoning. 
As shown in Fig.~\ref{fig:stereo_comparison}, FAST remains robust in challenging Booster scenes where both specialized methods exhibit noticeable local failures. 
In the first example, FAST recovers a spatially coherent disparity field with substantially lower errors. 
For the transparent object, S$^2$M$^2$ and FoundationStereo suffer from background-disparity leakage through the surface, whereas FAST produces a smoother and more coherent disparity estimate with fewer errors inside the transparent region. 
These results demonstrate that the generic correspondence formulation of FAST transfers effectively to challenging stereo scenes and provides complementary robustness beyond stereo-specific designs.

\subsection{Wide-baseline Dense Matching}
\begin{figure}[t]
  \centering
  \includegraphics[width=0.45\textwidth]{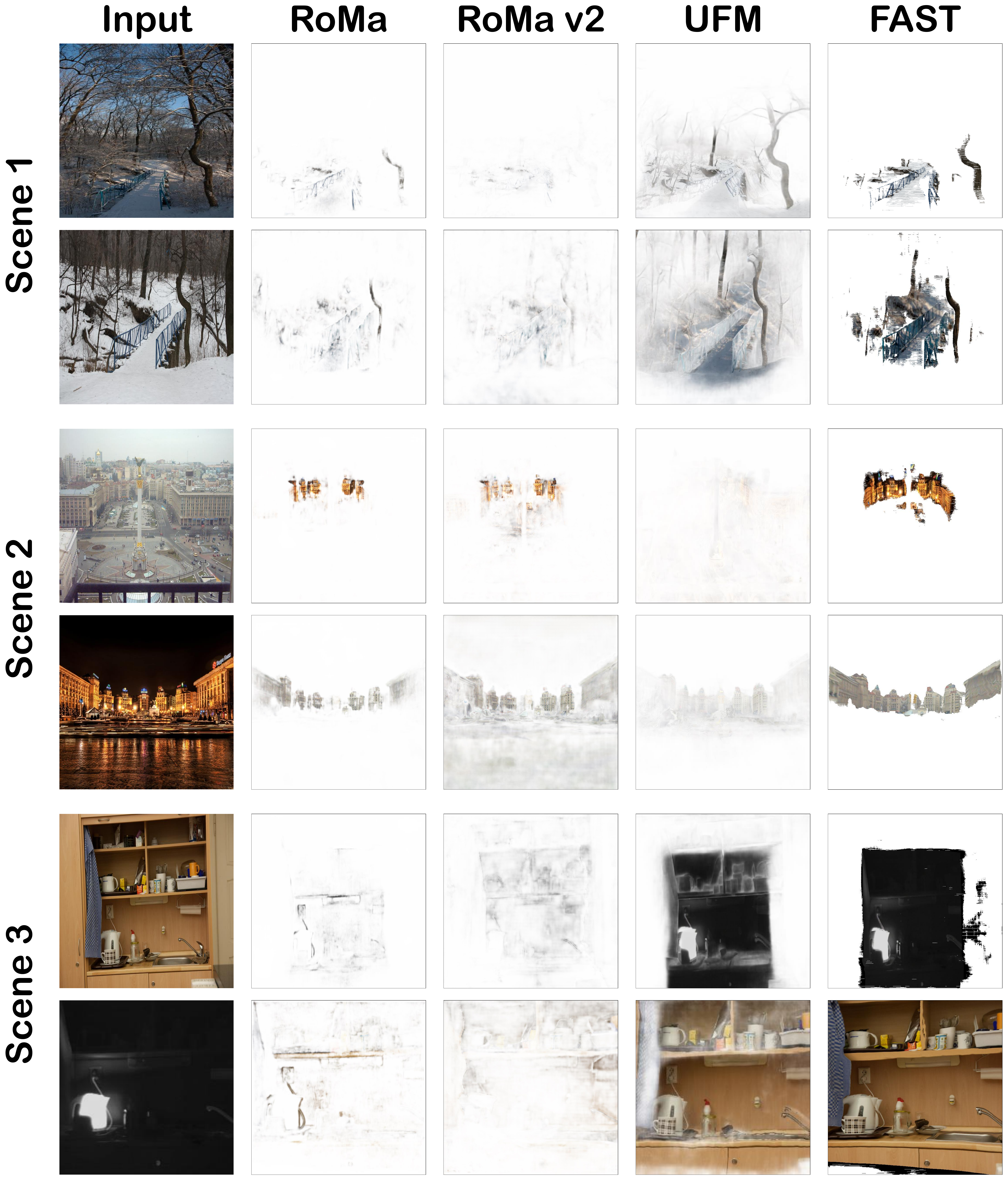}
  \caption{Dense matching comparison on the challenging WxBS scenes.
  }
  \label{fig:dense_matching_comparison}
\end{figure}

% 绘制两类图, 一个是给定 query point, 在另一个视图中找出对应匹配点 / 画出所有匹配线
% 第二类图绘制 warping 图像进行对比
We further evaluate FAST on WxBS~\cite{WxBS} against RoMa~\cite{RoMa}, RoMa v2~\cite{RoMav2}, and UFM~\cite{UFM}, three recent state-of-the-art dense correspondence methods. As shown in Fig.~\ref{fig:dense_matching_comparison}, the evaluated pairs cover extreme appearance variations, including seasonal changes, day--night transitions, and cross-modal RGB--infrared matching. RoMa and RoMa v2 often produce sparse warps with large low-confidence regions, while UFM provides denser predictions but fails to recover reliable correspondences in Scene 2. In contrast, FAST consistently produces more complete and structurally coherent warps with fewer spurious correspondences. The advantage is particularly pronounced for RGB--infrared matching, where FAST preserves most of the scene structure despite the substantial modality gap. These results demonstrate a favorable balance between correspondence completeness and outlier suppression, highlighting the generalization capability of FAST beyond conventional flow and stereo settings.

\begin{table*}[t]
  \centering
  \caption{
  Overview of the training data collection. The datasets are grouped into three categories: dynamic scenes with moving objects, static rigid scenes with small baselines, and static rigid scenes with wide baselines.
  For video datasets, (Seq, Stride) denotes that continuous frames are divided into clips of length \textsf{Seq} with a temporal stride of \textsf{Stride}.
  $^*$ indicates that optical flow labels are generated via a depth-to-flow strategy described in Appendix~\ref{sec:depth-to-flow}.
}
  \begin{tabular}{lccccccr}
  \toprule
  \multirow{2}{*}{Dataset} & \multirow{2}{*}{\begin{tabular}[c]{@{}c@{}}Dynamic\\ Scenes\end{tabular}} & \multicolumn{3}{c}{Lables}   & \multirow{2}{*}{Res.} & \multirow{2}{*}{(Seq, Stride)} & \multirow{2}{*}{Clips} \\
  \cmidrule(lr){3-5}
                           &                                                                           & Flow & Depth & Camera &                       &                                &                        \\
  \midrule
  AutoFlow~\cite{AutoFlow}                    & \cmark & \cmark & \xmark & \xmark & 448$\times$576               & (2,1)                          & 27295                  \\
  cvo~\cite{cvo}                              & \cmark & \cmark & \xmark & \xmark & 512$\times$512               & (2,1)                          & 125466                 \\
  DynamicReplica~\cite{dynamicreplica}        & \cmark & \cmark & \cmark & \cmark & 720$\times$1280              & (8,4)                          & 35742                  \\
  FlyingChairs~\cite{FlowNet}                 & \cmark & \cmark & \xmark & \xmark & 384$\times$512               & (2,1)                          & 22872                  \\
  Infinigen~\cite{InfiniGen}                  & \cmark & \cmark & \cmark & \cmark & 720$\times$1280              & (2,1)                          & 1910                   \\
  Kubric~\cite{Kubric}                        & \cmark & \cmark & \cmark & \cmark & 512$\times$512               & (2,1)                          & 135332                 \\
  SceneFlow~\cite{SceneFlow}                  & \cmark & \cmark & \cmark & \cmark & 540$\times$960               & (2,1)                          & 66252                  \\
  Sintel~\cite{sintel}                        & \cmark & \cmark & \xmark & \xmark & 436$\times$1024              & (2,1)                          & 3054                   \\
  Spring~\cite{Spring}                        & \cmark & \cmark & \cmark & \cmark & 1080$\times$1920             & (2,1)                          & 4889                   \\
  VirtualKITTI2~\cite{VirtualKITTI2}          & \cmark & \cmark & \cmark & \cmark & 375$\times$1242              & (2,1)                          & 21110                  \\
  BlinkVision~\cite{blinkvision}              & \cmark & \cmark & \cmark & \cmark & 540$\times$960               & (2,1)                          & 181960                  \\ \midrule
  EDEN~\cite{EDEN}                            & \xmark & \xmark & \cmark & \cmark & 480$\times$640               & (2,1)                          & $^*$364988             \\
  GTA5-SfM~\cite{GTA5-SfM}                    & \xmark & \xmark & \cmark & \cmark & 480$\times$640               & (2,1)                          & $^*$18358              \\
  MatrixCity~\cite{MatrixCity}                & \xmark & \xmark & \cmark & \cmark & 1000$\times$1000             & (2,1)                          & $^*$338979             \\
  Replica~\cite{replica}                      & \xmark & \xmark & \cmark & \cmark & 680$\times$1200              & (8,4)                          & $^*$3992               \\
  TartanAir~\cite{TartanAir}                  & \xmark & \cmark & \cmark & \cmark & 480$\times$640               & (2,1)                          & 305899                 \\
  TaitanAirv2~\cite{TartanAirv2}              & \xmark & \cmark & \cmark & \cmark & 480$\times$640               & (2,1)                          & 1424947                 \\ 
  CREStereo~\cite{crestereo}                  & \xmark & \cmark & \xmark & \xmark & 1080$\times$1920             & (2,1)                          & 200000                 \\
  FallingThings~\cite{FallingThings}          & \xmark & \xmark & \cmark & \cmark & 540$\times$960               & (2,1)                          & 61500                  \\
  FSD~\cite{FoundationStereo}                 & \xmark & \cmark & \xmark & \xmark & 720$\times$1280              & (2,1)                          & 1108890                \\
  KenBurns~\cite{KenBurns}                    & \xmark & \xmark & \cmark & \cmark & 512$\times$512               & (2,1)                          & 76048                  \\
  WMGStereo~\cite{WMGStereo}                  & \xmark & \cmark & \xmark & \cmark & 720$\times$1280              & (2,1)                          & 155729                 \\
  \midrule
  Hypersim~\cite{hypersim}                    & \xmark & \xmark & \cmark & \cmark & 768$\times$1024              & (2,1)                          & $^*$87374             \\
  Structure3D~\cite{Structured3D}             & \xmark & \xmark & \cmark & \cmark & 720$\times$1280              & (2,1)                          & $^*$82182             \\
  MegaDepth~\cite{MegaDepth}                  & \xmark & \xmark & \cmark & \cmark & 480$\times$640                & (2,1)                          & $^*$100150             \\
  BlendedMVS~\cite{BlendedMVS}                & \xmark & \xmark & \cmark & \cmark & 480$\times$640                & (2,1)                          & $^*$446175             \\
  TA-WB~\cite{UFM}                            & \xmark & \cmark & \xmark & \xmark & 480$\times$640               & (2,1)                          & $^*$279183             \\
  
  \bottomrule
  Flow-6M &&&&&&&5,680,276\\
  \bottomrule
  \end{tabular}
  \label{tab:data_collection}
  \vspace{-0.2cm}
\end{table*}

\subsection{Applications on 3D Reconstruction}

\begin{figure}[t]
  \centering
  \includegraphics[width=0.45\textwidth]{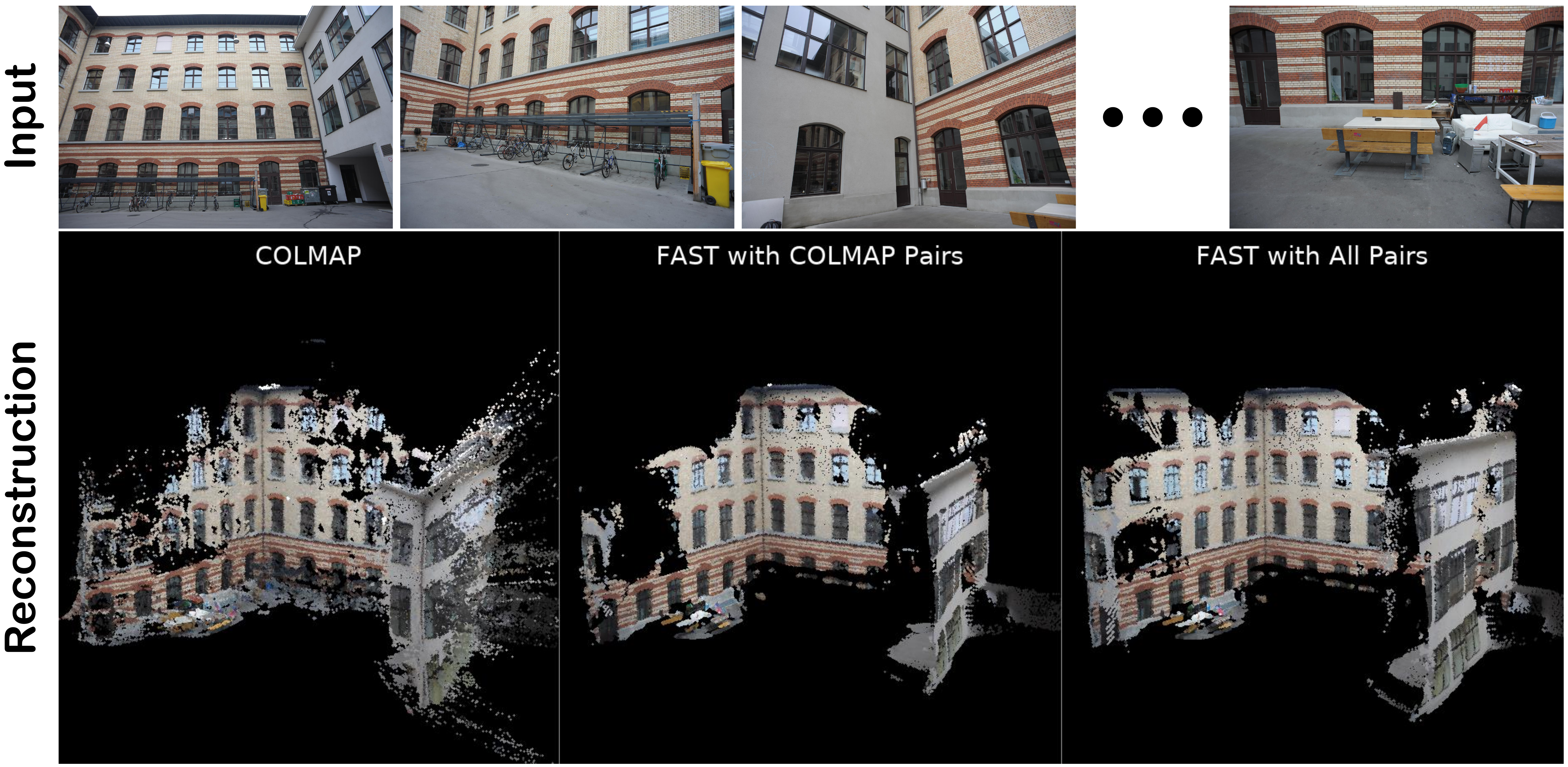}
  \caption{3D reconstruction of the ETH3D~\cite{eth3d} courtyard scene (38 views). From left to right: raw COLMAP, FAST COLMAP with COLMAP-computed pairs, and FAST with all possible pairs.
  }
  \label{fig:reconstruction_for_contrary}
\end{figure}

\begin{figure}[t]
  \centering
  \includegraphics[width=0.45\textwidth]{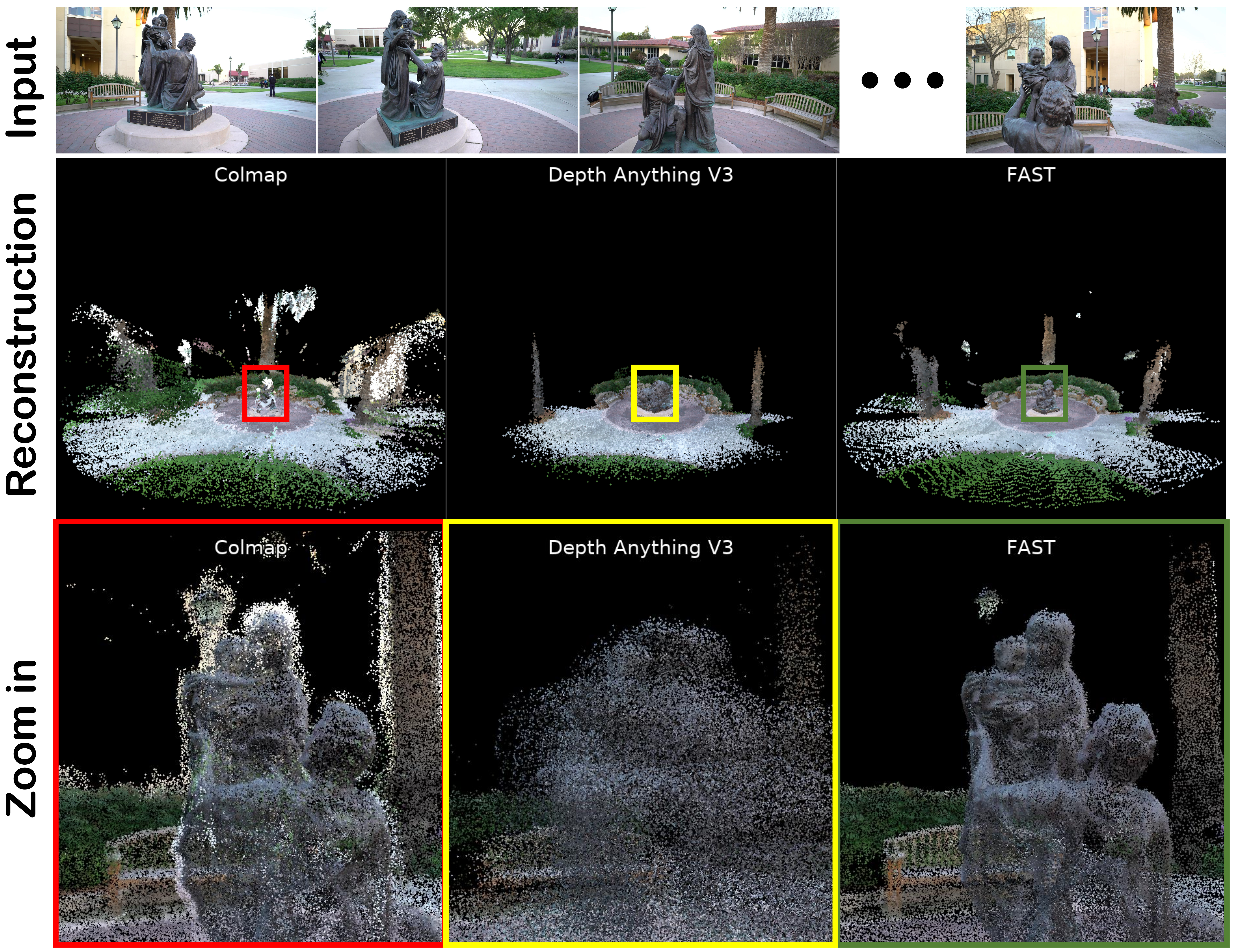}
  \caption{3D reconstruction of the T\&T~\cite{TandT} Family scene (152 views). From left to right: raw COLMAP, Depth Anything v3, and FAST COLMAP with COLMAP-computed pairs. 
  }
  \label{fig:reconstruction_for_family}
\end{figure}

We further evaluate FAST as a correspondence and depth-estimation component in a complete 3D reconstruction pipeline following~\cite{PanMatch}. 
To isolate the effect of FAST from image retrieval, we first compare it with the standard COLMAP pipeline under an identical image-pair graph. Only the pair identities are shared: COLMAP uses SIFT correspondences followed by PatchMatch stereo~\cite{patchmatch}, whereas FAST independently predicts correspondences, estimates camera poses through SfM and bundle adjustment, triangulates dense depth from flow, and produces the final fused point cloud. We additionally evaluate an exhaustive FAST variant that removes the COLMAP pairing prior and processes every unordered image pair. As shown on the ETH3D Courtyard scene (Fig.~\ref{fig:reconstruction_for_contrary}), the standard COLMAP reconstruction contains a substantial number of flying points, whereas both FAST variants produce visibly cleaner geometry. FAST with exhaustive pairing further improves point-cloud density and scene completeness over the variant restricted to the COLMAP-derived pair graph. This indicates that the dense correspondences predicted by FAST can exploit additional cross-view constraints that are not retained by the conventional pair-selection stage.

On the T\&T Family scene with long trajectory (Fig.~\ref{fig:reconstruction_for_family}), FAST-COLMAP again produces substantially fewer floating artifacts than standard COLMAP. Compared with recent feed-forward reconstruction approach like DA3~\cite{depthanything3}, FAST recovers finer geometric structures, particularly the silhouette and local details of the human figure, while also preserving more complete scene coverage. DA3 reconstructs the dominant scene structure but tends to produce smoother and less detailed geometry. These qualitative results demonstrate that FAST can provide accurate correspondences for both camera-pose recovery and dense reconstruction, yielding clean and detailed point clouds without relying on COLMAP pose or depth estimates.
%-----------------------------------------------------------------------------

\section{Data Collection}
Details of our training datasets, including image resolutions, ground-truth labels, and sample scales, are summarized in Table~\ref{tab:data_collection}.

%--------------------------------------------------------------------------------

% \ifCLASSOPTIONcompsoc
%   % The Computer Society usually uses the plural form
%   \section*{Acknowledgments}
% \else
%   % regular IEEE prefers the singular form
% \fi


\ifCLASSOPTIONcaptionsoff
  \newpage
\fi



\bibliographystyle{IEEEtran}
\footnotesize{
\bibliography{main}
}

\end{document}